\documentclass[journal]{IEEEtran}
\usepackage{hyperref} 
\usepackage{url}
\usepackage{booktabs}
\usepackage{longtable} 
\usepackage{pifont}
\newcommand{\cmark}{\ding{51}}%
\newcommand{\xmark}{\ding{55}}%
\usepackage{booktabs}
\usepackage{xcolor}

\usepackage{comment}
\usepackage{dblfloatfix}
\usepackage{afterpage}
\usepackage[utf8]{inputenc}
\usepackage{cite}
\usepackage{amsmath}
\usepackage{array}
\usepackage{tabularx}
\usepackage{graphicx}
\usepackage{multirow}

\ifCLASSINFOpdf
\else
 
\fi

\title{A Comprehensive Survey of Foundation Models in Medicine}

\author{Wasif~Khan\textsuperscript{a},
        Seowung Leem\textsuperscript{a},
        Kyle B. See\textsuperscript{a},
        Joshua K. Wong\textsuperscript{b,c},
        Shaoting Zhang\textsuperscript{d}, and
        Ruogu Fang\textsuperscript{a,e,f,g}*
        
\thanks{a. J. Crayton Pruitt Family Department of Biomedical Engineering, Herbert Wertheim College of Engineering, University of Florida, Gainesville, FL. USA.\\
b. Norman Fixel Institute for Neurological Diseases, University of Florida, Gainesville, FL, USA.\\
c. Department of Neurology, University of Florida, Gainesville, FL, USA.\\
d. Department of Computer Science, the University of North Carolina at Charlotte, Charlotte, NC, USA\\
e. Center for Cognitive Aging and Memory, McKnight Brain Institute, University of Florida, Gainesville, FL. USA.\\
f. Department of Electrical and Computer Engineering, Herbert Wertheim College of Engineering, University of Florida, Gainesville, FL. USA.\\
g. Department of Computer Information Science Engineering, Herbert Wertheim College of Engineering, University of Florida, Gainesville, FL. USA.\\
* Corresponding author: Ruogu Fang, ruogu.fang@bme.ufl.edu}
\thanks{Manuscript received XX; revised August XX.}}

\begin{document}

\maketitle

\begin{abstract}
Foundation models (FMs) are large-scale deep learning models trained on massive datasets, often using self-supervised learning techniques. These models serve as a versatile base for a wide range of downstream tasks, including those in medicine and healthcare. FMs have demonstrated remarkable success across multiple healthcare domains. 
However, existing surveys in this field do not comprehensively cover all areas where FMs have made significant strides. In this survey, we present a comprehensive review of FMs in medicine, focusing on their evolution, learning strategies, flagship models, applications, and associated challenges. We examine how prominent FMs, such as the BERT and GPT families, are transforming various aspects of healthcare, including clinical large language models, medical image analysis, and omics research. Additionally, we provide a detailed taxonomy of FM-enabled healthcare applications, spanning clinical natural language processing, medical computer vision, graph learning, and other biology- and omics- related tasks. Despite the transformative potentials of FMs, they also pose unique challenges. This survey delves into these challenges and highlights open research questions and lessons learned to guide researchers and practitioners. Our goal is to provide valuable insights into the capabilities of FMs in health, facilitating responsible deployment and mitigating associated risks. 
\end{abstract}


\IEEEpeerreviewmaketitle

\section{Introduction}
\subsection{History}

\IEEEPARstart{A}{rtificial} intelligence (AI) has reached significant milestones in recent decades. This progress is mainly attributed to the enhanced availability of large-scale data, computational resources, and intelligent algorithm design \cite{ramesh2004artificial}. AI-based techniques have been used for decades within healthcare \cite{panesar2019machine}. For example, computer-assisted surgery began in the early 1980s, which marked the start of medical AI \cite{kwoh1988robot}. In 1989, a small convolutional neural network (CNN) model demonstrated promising performance by using a backpropagation algorithm to recognize handwritten digits \cite{lecun1989backpropagation}. However, the application of AI in medicine was not facilitated until the AI spring, which was initiated by the emergence of the AlexNet model \cite{krizhevsky2012imagenet}. The significant momentum in the evolution of AI models was demonstrated after the arrival of the AlexNet model. Subsequently, 
models such as VGGNet and ResNet further advanced the field of image classification, object detection, and segmentation.
Following CNN, Recurrent neural networks (RNN) \cite{hochreiter1997long} were proposed for sequential data processing that showed promising performance in various domains, including time series analysis, natural language processing (NLP), and speech recognition \cite{lalapura2021recurrent}. Another significant contribution came with the introduction of generative adversarial networks (GAN) \cite{goodfellow2014generative} for generating synthetic data. 

Transformers, introduced in 2017, formed a crucial foundation of the most powerful deep learning models available today. Transformers introduced a self-attention mechanism to capture sequences in parallel and addressed the limitations of computational and sequential RNNs. 
Initially designed for NLP, transformers have been extended to other fields, including computer vision such as Vision Transformer (ViT), introduced in 2021 \cite{ViT}, which is tailored specifically for vision tasks.
Transformer models have led to the development of foundation models (FM) \cite{bommasani2021opportunities}, which are large-scale pre-trained models trained on vast amounts of data and fine-tuned to perform a wide array of downstream tasks. The key enablers of FMs include training data, base model, transfer learning, and scale \cite{bommasani2021opportunities}. FMs adopt a self-supervised learning (SSL) approach, unlike traditional DL models that are dependent on large amounts of data. In this approach, the pre-training task is autonomously generated from unlabeled data. SSL evolved from word embeddings that provided context-independent vectors for words to autoregressive language modeling, which predicts the subsequent word based on previous words \cite{howard2018universal,gpt}. 

SSL led to one of the first FMs known as BERT \cite{devlin2018bert}, a powerful transformer architecture characterized by its bidirectional sentence encoders and scalability extending to larger architectures and datasets. BERT became the standard model in 2019. Many subsequent models were developed based on BERT architectures, including RoBERTa \cite{liu2019roberta}, BART \cite{lewis2019bart}, and T5 \cite{raffel2020exploring}. These models have become prevalent in the development of state-of-the-art (SOTA) NLP models. These NLP models are termed large language models (LLM). Despite being available since late 2018, these models gained broader recognition across various domains following the introduction of ChatGPT in late 2022\footnote{\url{https://chat.openai.com/}}. 

Similarly, vision-based models such as ViT, Swin Transformers \cite{swinT}, and Data-efficient Image Transformer \cite{touvron2021training} have demonstrated SOTA performance in visual recognition tasks. Once pre-trained, they can be fine-tuned for various healthcare applications. Furthermore, these models were soon generalized to different fields, such as text \cite{kim2023chatgpt}, videos \cite{zhou2024survey}, speech \cite{zaver2023opportunities}, tabular data \cite{yin2020tabert}, protein sequences \cite{rives2021biological}, organic molecules \cite{rothchild2021c5t5}, and reinforcement learning \cite{cao2024survey}. 

Healthcare constitutes a substantial economic sector representing 17\% of the United States GDP \cite{bommasani2021opportunities}. Many specialized models have been developed for various healthcare tasks \cite{miotto2018deep}. However, they are based on limited labeled data and face challenges when generalized to different tasks. It remains challenging to adopt general-purpose FMs in the healthcare domain due to the complex structures and limited publicly available medical data. Therefore, several works have been carried out to define FMs specifically for medical data. These models are task-agnostic and can be generalized across different tasks within the medical domain. For instance, the GatorTronGPT \cite{GatorTrongpt} model was built on the GPT-3 \cite{gpt3} architecture and includes 20 billion parameters. It was trained on 277 billion words with 82 billion words of clinical text, which were obtained from 2 million patients. GatorTronGPT exhibits strong performance in various medical tasks such as named entity recognition (NER), biomedical question answering (QA), biomedical relation extraction (RE), natural language inference (NLI), and semantic textual similarity. Similarly, models such as MedSAM \cite{medsam}  trained on 1.5 million medical images are used for medical image segmentation and surpass other models \cite{UNET, SAM}. FMs designed for other tasks within the medical domain are described in detail in Section \ref{FMmedicine}. 

\subsection{Research Objectives and Outline}
\begin{figure}[t]
    \centering
    \includegraphics[width=\columnwidth]{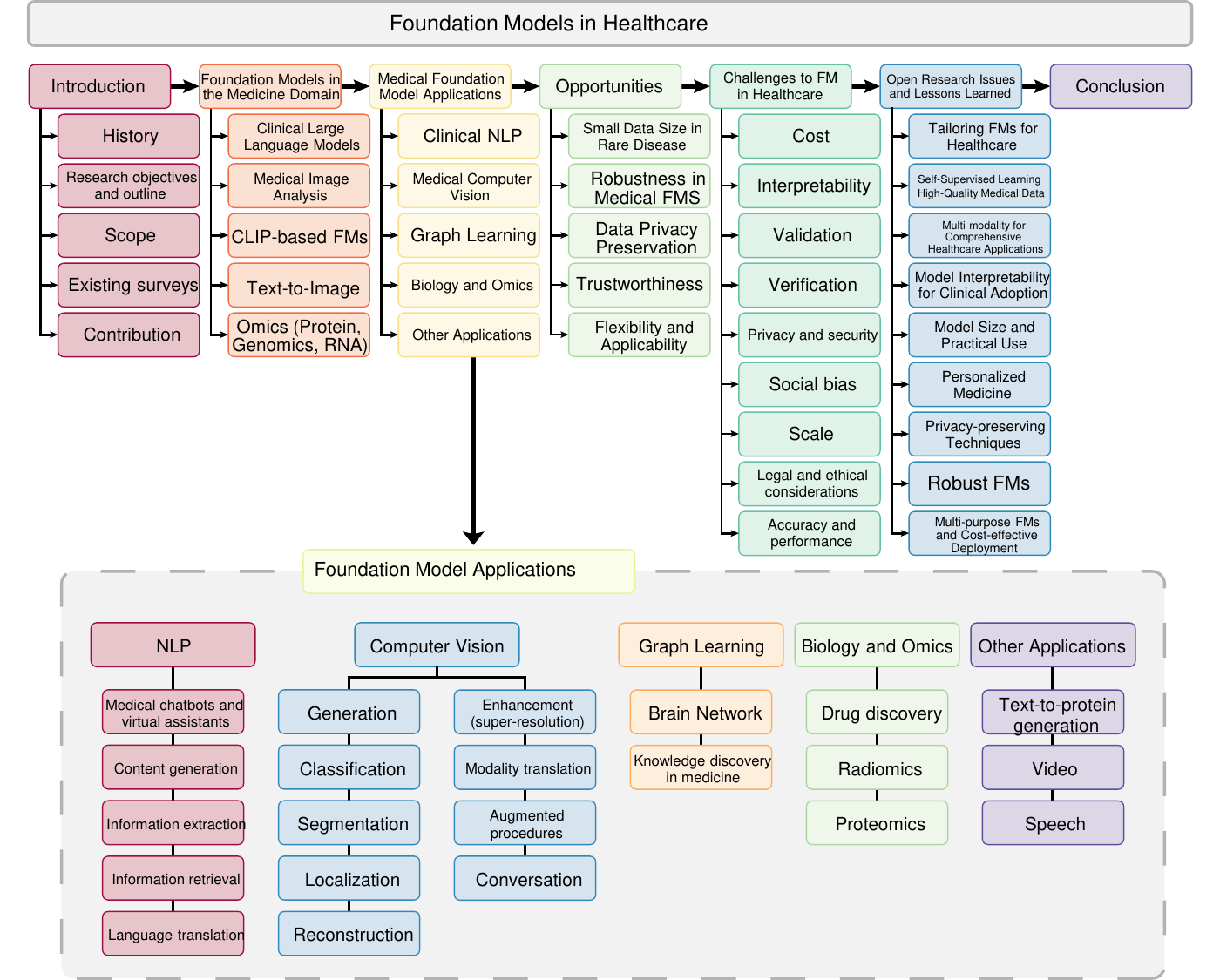}
    \caption{Outline of the survey paper. We present an overview of medical FMs, followed by their applications in healthcare. Additionally, we offer in-depth insights into the taxonomy of medical FMs, the opportunities they enable, the challenges they face, and open research questions and lessons learned.}
    \label{fig:fig1}
\end{figure}

This survey paper aims to comprehensively review FMs in the healthcare domain. First, we present an overview of the history of artificial intelligence (AI) models, followed by the development of transformers-based models (Figure \ref{fig:fig1}). The survey is structured to provide a comprehensive study of FMs, explaining the background of FMs (Supplementary Materials Section I), followed by an extensive overview of FMs (Supplementary Materials Section II). The flagship FMs such as the BERT family, GPT models, CLIP, stable diffusion, and industrially scaled pre-trained models, are explained in Supplementary Materials Section III). The core FMs specifically designed for medical applications are presented in Section~\ref{FMmedicine}. The applications of FMs in the healthcare domain are presented in Section~\ref{FMapplications}. We present multiple applications, including clinical NLP (Section~\ref{NLP}), medical computer vision (Section~\ref{computervision}), healthcare graph learning (Section~\ref{graphLearning}), biology and omics (Section~\ref{bioandomics}), and others (Section~\ref{others}). Section~\ref{Opportunities} explains the available opportunities associated with FMs, including controllability, adaptability, and applicability. The challenges associated with FMs, such as cost, interpretability, validation, and scale, are thoroughly discussed in Section~\ref{challenges}. We highlighted the open research question and potential lessons learned in Section~\ref{openresearch}, and a conclusion in Section~\ref{conclusion}.

\subsection{Scope}
FMs have been widely adopted across multiple domains. However, a detailed taxonomy covering all aspects of FMs within the medical domain is still missing. Existing surveys have concentrated on general-purpose FMs \cite{bommasani2021opportunities,zhou2023comprehensive,yang2023harnessing,awais2023foundational,kolides2023artificial}, whereas this survey focuses on the healthcare domain. Furthermore, we describe the major flagship models that led to the development of medical-based FMs (Supplementary Materials Section III).

\subsection{Existing surveys}
In 2023, the rapid advances in FMs led to the publication of multiple survey articles (Table~\ref{tab:comparison}). However, the majority of these articles do not focus on healthcare. Some articles only focus on LLMs on \cite{zhao2023survey}, which excluded FMs for other modalities such as diffusion models \cite{yang2023diffusion}. Furthermore, existing articles focused on healthcare lack a detailed taxonomy of FMs that includes all of its associated aspects, such as LLMs, vision, protein, audio, and graph FMs.

\begingroup
\footnotesize
\begin{table}[htbp]
    \caption{Comparison with existing survey.}
    \label{tab:comparison}
    \centering
    \setlength{\tabcolsep}{5pt} 
    \renewcommand{\arraystretch}{1.2} 
    \begin{tabular}{p{2.5cm} p{1.1cm} p{0.6cm} p{0.6cm} p{0.6cm} p{0.6cm} p{0.6cm}} 
        \toprule
        \textbf{Work} & \textbf{Healthcare} & \textbf{LLMs} & \textbf{Vision FMs} & \textbf{Protein} & \textbf{Audio} & \textbf{Graph FMs} \\
        \midrule
        Wornow et al. \cite{wornow2023shaky} & \cmark & \cmark & \xmark & \xmark & \xmark & \xmark \\
        Zhou et al.  \cite{zhou2023comprehensive} & \xmark & \cmark & \cmark & \xmark & \cmark & \cmark \\
        Yang et al. \cite{yang2023diffusion} & \xmark & \cmark & \cmark & \xmark & \xmark & \xmark \\
        Yang et al. \cite{yang2023harnessing} & \xmark & \cmark & \xmark & \xmark & \xmark & \xmark \\
        Awais et al. \cite{awais2023foundational} & \xmark & \xmark & \cmark & \xmark & \cmark & \xmark \\
        Bommasani et al.  \cite{bommasani2021opportunities} & \xmark & \cmark & \cmark & \cmark & \cmark & \cmark \\
        Kolides et al. \cite{kolides2023artificial} & \xmark & \cmark & \cmark & \xmark & \cmark & \xmark \\
        Hadi et al.  \cite{hadi2023large} & \xmark & \cmark & \cmark & \xmark & \cmark & \xmark \\
        Azad et al.  \cite{azad2023foundational} & \cmark & \xmark & \cmark & \xmark & \xmark & \xmark \\
        Zhou et al.  \cite{zhou2023survey} & \cmark & \cmark & \xmark & \xmark & \xmark & \xmark \\
        Zhao et al.  \cite{zhao2023clip} & \cmark & \xmark & \cmark & \xmark & \xmark & \xmark \\
        Li et al.  \cite{sun2024medical} & \cmark & \cmark & \cmark & \xmark & \xmark & \xmark \\
        This survey & \cmark & \cmark & \cmark & \cmark & \cmark & \cmark \\
        \bottomrule
    \end{tabular}
\end{table}
\endgroup

\subsection{Contribution}
This survey provides a detailed analysis of recent works published until early 2024. It encompasses the evolutionary journey of FMs in the healthcare sector, providing insights into their historical development and categorization. The survey explores the applications of FMs in various healthcare domains including clinical NLP, medical computer vision, biology and omics, and other relevant areas. Furthermore, we highlight the major challenges associated with FMs and outline future directions associated with FMs in healthcare.

\section{Foundation Models in the Medicine Domain}
\label{FMmedicine}
\begin{figure}[t]
    \centering
    \includegraphics[width=0.99\columnwidth]{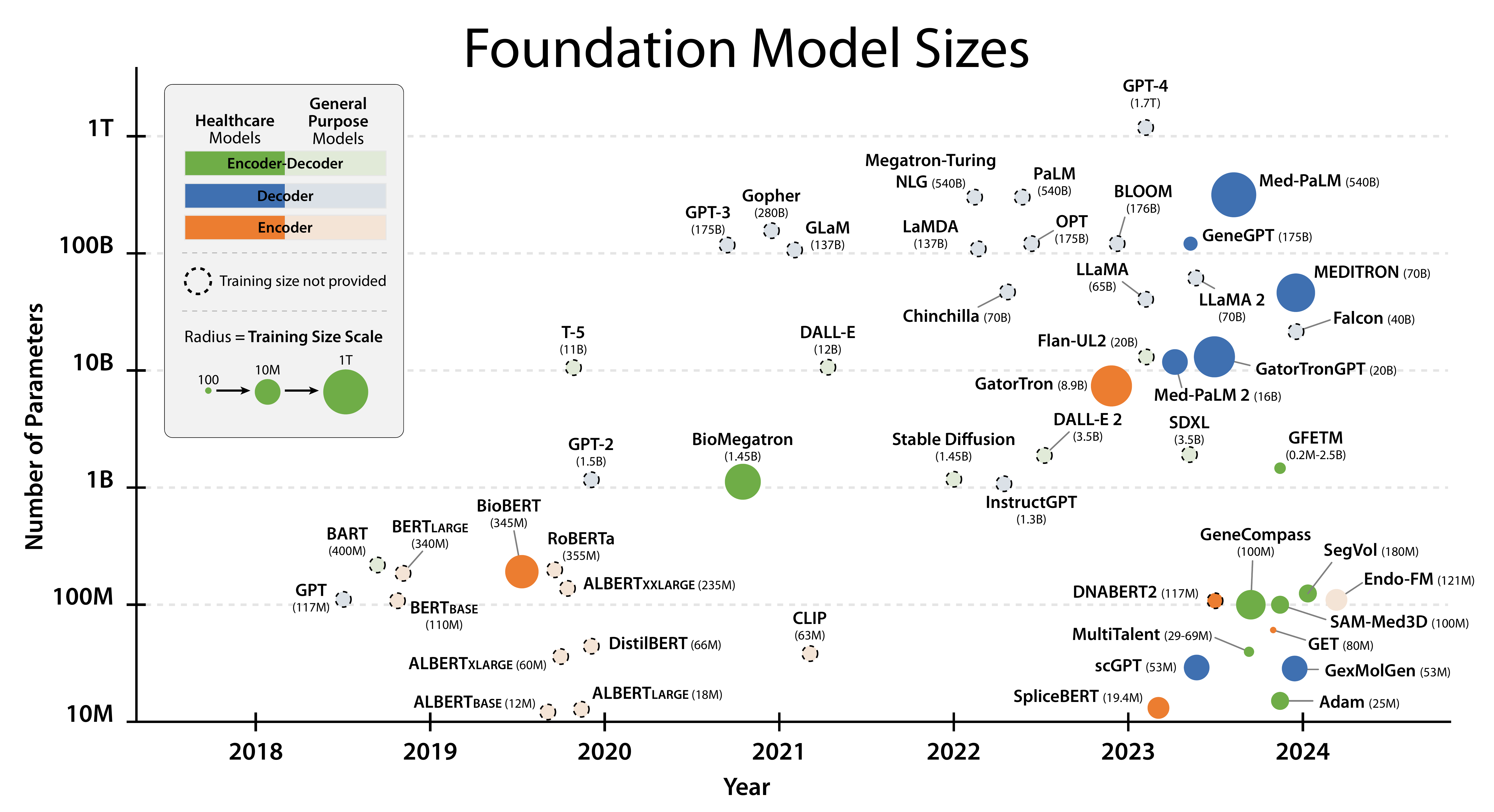}
    \caption{Learning architecture, model size, and training data used by representative foundation models. Details can be found in Supplementary Materials Section II.}
    \label{fig:fig2}
\end{figure}

Multiple FMs have been developed for general purposes that can be adopted across different domains. The main architectures are encoder, decoder, and encoder-decoder (Supplementary Materials Section-II A). The distribution of the representative models depending on their learning architectures is shown in Figure \ref{fig:fig2}. A significant number of FMs are based on architectures like BERT, GPT, and Stable Diffusion. Therefore, brief explanations of these models are provided in the Supplementary Materials (Section III).

Although multiple general-purpose FMs are developed, directly adopting them for healthcare could be challenging. For instance, models such as Word2Vec, ELMo, and BERT, primarily trained on general domain datasets, make their performance on biomedical texts uncertain due to significant differences in word distributions \cite{biobert}. Similarly, employing vision-text pre-training models like CLIP poses challenges because of its extensive data requirements and the subtle, fine-grained differences in medical images compared to general images \cite{biomedclip}. Furthermore, the diverse imaging modalities in medicine complicate the development of unified models, as seen with the Segment Anything Model (SAM), which is ineffective for 3D volumetric medical images due to its 2D architecture \cite{Sammed3d}. Therefore, there is a need for specialized FMs tailored to the unique demands of the healthcare domain. These emerging models have shown to be highly successful (Table \ref{tab:Flagship}). This section highlights FMs exclusively designed for healthcare applications that tackle challenges and harness opportunities within the healthcare domain.



\subsection{Clinical Large Language Models (CLLM)}
\label{CLLM}

BioBERT is one of the earliest domain-specific LLMs \cite{biobert}. It is a biomedical language processing model that was trained on large-scale biomedical corpora. BioBERT demonstrated impressive performance across 15 datasets addressing three NLP tasks: NER, RE, and QA. Another model called BioMegatron \cite{biomegatron} employs a BERT-style architecture with up to 1.2 billion parameters.
It is trained on a larger domain corpus (PubMed abstracts and PMC full-text) that outperforms previous SOTA models on biomedical NLP benchmarks (NER, RE, QAs).
GatorTron \cite{GatorTron} is another SOTA large-scale clinical language model developed from scratch was designed for NLP tasks on EHRs. It has 8.9 billion parameters and is trained on a corpus of over 90 billion words. These data included substantial clinical text from the University of Florida (UF) Health, PubMed, Wikipedia, and MIMIC III \cite{mimicIII}. It outperforms current biomedical and clinical transformers across five distinct clinical NLP tasks.
ChatGPT is a general-purpose model trained on various datasets, including a portion of healthcare data. However, the primary focus is not limited to the healthcare domain. Inspired by ChatGPT, GatorTron \cite{GatorTron} was further extended to the GatorTronGPT \cite{GatorTrongpt} model. This was built on the GPT-3 architecture with 20 billion parameters. The training corpus included 277 billion words with 195 billion words from the Pile dataset and 82 billion words from de-identified clinical text obtained from UF Health. GatorTronGPT exhibits a remarkable ability to generate synthetic data (GatorTronS) aligned with medical knowledge, as validated through various clinical tasks. Furthermore, the performance on different tasks including NER, QA, RE, NLI, and semantic textual similarity achieved promising performance. Moreover, it has been tested on the physicians' Turing test and achieved comparable performance to humans. Singhal et al. \cite{medpalm} proposed Med-PaLM, evaluated on a well-curated MultiMedQA benchmark. The proposed model achieved promising performance; however, it was slightly inferior compared to clinicians. Another model built upon PaLM-2 (Med-PaLM-2) was fine-tuned on the MultiMedQA benchmark and achieved 19\% better performance than Med-PaLM. The more recent MEDITRON model \cite{meditron} is a large-scale medical language model with 70 billion parameters trained with 48.1 billion tokens from the medical domain. It has outperformed previous SOTA models including GPT-3.5 and MedPaLM.

\subsection{Medical Image Analysis}
\label{MIA}

FMs are trained on large numbers of training images, enabling fine-tuning for diverse tasks. However, the patterns and features in medical images are different from those in natural images \cite{zhang2023challenges}. Multiple works have been done for robust domain-specific models for medical images. For instance, the Segment Anything Model (SAM) \cite{SAM} was originally developed for general-purpose segmentation and trained on over 11 million images. It has been fine-tuned for medical image applications with great success shown by \cite{deng2023segment}. A medical-related FM for universal medical image segmentation called MedSAM was proposed by Ma et al. \cite{medsam} which employs a vision transformer-based (ViT) image encoder for feature extraction and a mask decoder for segmentation results. MedSAM uses multiple online medical datasets with 1,570,263 images and text pairs for training. A detailed evaluation showed promising performance compared to other segmentation models, including SAM \cite{SAM} and U-Net \cite{UNET}. Following the success of MedSAM, numerous models have emerged for medical image segmentation \cite{du2023segvol, Sammed3d, zhang2024segment}. The details of these segmentation tasks are explained in Section~\ref{segmentation}.

\subsection{CLIP-based FMs}
\label{CLIP}
CLIP \cite{CLIP} is a robust neural network architecture with zero-shot capabilities \cite{gpt2} that achieves various classification benchmarks using natural language instructions, without direct optimization for each specific benchmark \cite{gpt2}. It is a simplified variant of ConVIRT \cite{zhang2022contrastive} and is trained on a dataset called WebImageText, comprising 400 million pairs of images and corresponding text collected. 
Although CLIP is a robust model trained on large amounts of data, medical image-text datasets are of a much smaller scale than general internet data. Therefore, CLIP can be prone to false negatives, such as incorrectly treating images and reports from separate patients as dissimilar, despite their shared semantic content. Furthermore, it relies on image-text pairs while medical images are provided with medical domain labels compared to raw text. Therefore, Wang et al. \cite{wang2022medclip} proposed MedCLIP, which decouples images and texts for contrastive learning, significantly increasing the usable training data and replacing the InfoNCE loss \cite{oord2018representation} with a semantic matching loss based on medical knowledge (semantic similarity) to eliminate false negatives. BioClinicalBERT \cite{alsentzer2019publicly} \footnote{\url{https://huggingface.co/emilyalsentzer/Bio_ClinicalBERT}} has been used as the text encoder and Swin Transformer \cite{swinT} or ResNet-50 \cite{resnet} have been used as the vision encoder. Detailed experimental results showed that the proposed MedCLIP achieved promising performance compared to CLIP and other methods \cite{gloria,zhang2022contrastive} with 10 times fewer training data.
Zhang et al. \cite{biomedclip} proposed BiomedCLIP, which outperformed the original CLIP in the biomedical domain. To address the growing need for large-scale datasets in the field of biomedical research, Lin et al. \cite{pmcclip} introduced a substantial dataset comprising more than 1.6 million image-text pairs. The dataset was collected from over 2.4 million scientific papers with over 381K medical figure-caption pairs. Liu et al. \cite{liu2023clip} proposed a CLIP-based framework that uses text embeddings from CLIP to enhance segmentation models for automated organ segmentation and tumor detection which can efficiently segment 25 organs. However, their proposed method relies significantly on one labeled training dataset. Therefore, Liu et al. \cite{liu2023clip} proposed a CLIP-based universal model that incorporates text embeddings and masked backpropagation.

\subsection{Text-to-Image}
\label{testtoimage}
 
Text-to-image models aim to generate visually realistic images from textual descriptions and have emerged as one of the most impressive applications \cite{zhang2023textS} in the era of FM. GAN-based works can generate images from text ( \cite{zhang2023textS}). However, these models tend to suffer from a lack of generalization, mode collapse, diversity, and delicacy in text-to-image details. 
FMs such as DALL-E \cite{DALLE} can generate images for various textual descriptions. It also exhibits the ability to modify the attributes of objects and control the number of instances present in the generated images. Similarly, another model called Pathways Autoregressive Text-to-Image (Parti) \cite{yu2022scaling} produces high-fidelity photorealistic images with complex compositions and rich content synthesis. However, to reduce the computational costs and sequential error accumulation \cite{zhang2023textS}, diffusion models \cite{ho2020denoising, stabledifuusion, sdxl} have emerged as a promising alternative.

Text-to-image is also beneficial in healthcare such as work done by Kim et al. \cite{kim2024controllable} who proposed FMs to generate text-conditional magnetic resonance imaging (MRI) scans. Similarly, Ali et al. \cite{ali2024demographic} compared DALL-E 2, Midjourney, and Stable Diffusion to represent demographic disparities in the surgical profession. 
Yang et al. \cite{yang2023medxchat} proposed MedXChat for Text-to-CXR synthesis. The proposed MedXChat outperforms existing models in generating accurate CXR images and reports, demonstrating superior adaptability across tasks. Similarly, a work done by Woolsey et al. \cite{woolsey2024diffusion} examines how text-to-image models can enhance health literacy by improving medical text comprehension. Synthesizing high-quality medical images assists physicians in better diagnosing complex cases and reducing the need for repeated scans, lowering radiation exposure for patients \cite{shi2024diffusion, lim2024exploring}.

\subsection{Omics}
\label{omics}
Human cells are basic structural and functional units that contain different entities such as proteins, nucleic acids (DNA, RNA), carbohydrates, and lipids \cite{hao2023large}. Multiple ML models have been used to extract useful patterns from the complex data structures associated with various omics domains, which include genomics, proteinomics, metabolomics, transcriptomics, Lipidomics, microbiomics, etc. \cite{reel2021using}. However, due to a lack of model generalizability, FMs are now widely used for omics data. For instance, Yang et al. \cite{scbert} proposed a BERT-based approach named scBERT for Cell type annotation. The encoder of the BERT model cannot be scaled to more than 512 gene inputs, so it was replaced with a performer \cite{performers} that can be scaled to 1600 gene inputs. Thus, scBERT does not depend on dimensionality reduction techniques and is capable of the full gene-level interpretation. scBERT performed better than existing models when evaluated on various tasks from seven scRNA-seq datasets.
A BERT-based model for RNA named RNABERT \cite{rnabert} was proposed by Akiyama for ncRNA sequence analysis, focusing on structural alignment and clustering. 
Chen et al. \cite{rnaFM} showed that most of the existing models rely on annotated RNA sequences which results in small training data (less than 30K samples) \cite{rnaFM} and omits large amounts of unannotated data. Therefore, they proposed a BERT-based RNA-FM model that utilizes 23 million unannotated RNA sequences in a self-supervised manner. However, Zhang et al. \cite{rnaMSM} showed that this approach struggles to capture evolutionary information from less conserved RNA sequences. Therefore, they proposed an RNA-MSM model \cite{rnaMSM}  based on multiple sequence alignment using homologous sequences from the RNAcmap pipeline \cite{zhang2021rnacmap}. 
Furthermore, a model named BigRNA \cite{bigRNA} also utilizes RNA-seq data obtained from GTEx consortium V6 release \cite{gtex2020gtex} that can be utilized for various tasks for RNA biology. Chen et al. proposed SpliceBERT \cite{spliceBERT} for sequence-based RNA splicing prediction. It was trained on pre-mRNA (precursor messenger RNA) sequences from 72 vertebrates, creating a pre-training dataset of approximately 2 million sequences. A more recent model, scHyena, was proposed by Oh et al \cite{schyena} for scRNA-seq analysis data in the brain. It outperforms existing methods in various tasks such as cell type classification and imputing biologically relevant values in scRNA-seq data.

Theodoris et al. \cite{geneformer} proposed the transformer-based Geneformer, a context-aware model to map the gene networks. The Geneformer \cite{geneformer} architecture consists of six encoder units, each with self-attention layers and a feedforward neural network, and is trained on 29.9 million single-cell transcriptomes (Genecorpus-30M collected from publicly available datasets). Cui \cite{scgpt} proposed the scGPT model trained on more than 33 million cells (from publicly available datasets) for single-cell biology applications. 
Yang et al. \cite{genecompass} developed GeneCompass, a cross-species transformer FM containing 120 million single-cell transcriptomes. Chen et al. \cite{genept} showed that although these models \cite{geneformer, scgpt,scbert} achieved better performance, they heavily rely on well-curated datasets and are computationally expensive. Therefore, to address the challenges of content hallucination and answer genomics questions, GeneGPT was proposed, which utilizes ChatGPT embeddings of different genes from the literature. The authors created embeddings using GPT-3.5 from data of 33,000 genes obtained from Geneformer \cite{geneformer} and scGPT \cite{scgpt}.
DNABERT \cite{dnabert} was one of the first few BERT-based models to decode the language of non-coding DNA by developing a comprehensive and transferable understanding of genomic DNA sequences. An extension, DNABERT2 \cite{dnabert2}, replaced the k-mer tokenization with byte pair encoding. It achieved promising performance with 21 times fewer parameters than DNABERT \cite{dnabert}.
Cheng et al. \cite{GexMolGen} proposed the scGPT-based Gene Expression-based Molecule Generator (GexMolGen) for designing hit-like molecules utilizing gene expression signatures. This model addresses the limitation of traditional similarity searching-based approaches, which search within a limited dataset. GexMolGen generates molecules capable of inducing the required transcriptome profile, exhibiting high similarity to known gene inhibitors.
Fu et al. \cite{getModel} proposed an interpretable FM named the General Expression Transformer (GET) for transcription regulation cell types. It demonstrates experimental-level accuracy in predicting gene expression, including in cell types not previously observed. Fan et al. \cite{gfetm} proposed a Genome Foundation Embedded Topic Model (GFETM) for single-cell ATAC-seq (scATAC-seq) data. 
Another model, scBERT \cite{scbert}, addresses the challenges in annotating cell types using single-cell RNA-seq data. The scBERT model leverages pretraining on large unlabelled datasets and fine-tuning on specific data to achieve robust cell type annotation, novel cell type discovery, and interoperability. Other works include \cite{rosen2023universal, tangbuilding, luo2023molfm, lu2023towards, hao2023large, israel2023foundation, franke2023scalable}. 
Alsabbagh et al. \cite{alsabbagh2023foundation} evaluated the performance of scGPT \cite{scgpt}, scBERT \cite{scbert}, and Geneformer \cite{geneformer} on an imbalance cell type dataset. The results on Zheng68K and Multiple Sclerosis datasets show that scGPT \cite{scgpt} and scBERT \cite{scbert} achieved better performance, but the Geneformer \cite{geneformer} performed the worst with 20\% lower accuracy. 


\section{Medical Foundation Model Applications}
\label{FMapplications}

\subsection{Clinical NLP}
\label{NLP}

\subsubsection{Medical Chatbots and virtual assistants}
FMs have advanced the use of chatbots in the medical domain. These medical chatbots are mainly based on BERT architecture and GPT-based models. For instance, Hamidi and Roberts \cite{hamidi2023evaluation} evaluated the performance of ChatGPT and Claude for patient-specific QA from clinical notes and showed that they provided accurate and relevant answers. Furthermore, a study by Shahsavar and Choudhury \cite{shahsavar2023user} showed that the majority (78.4\%) of participants (607 in total) were willing to use ChatGPT for self-diagnosis in healthcare. Similarly, medical chatbots can provide recommendations of comparable quality to those of human experts \cite{shahsavar2023user}. 
Sun et al. \cite{sun2023pathasst} proposed PathAsst, a generative AI assistant for diagnostic and predictive analytics in pathology. PathAssist was trained on a 195K dataset of pathology image-text pairs and employed a base model architecture with a CLIP vision encoder and Vicuna-13B LLM. A voice-based digital assistance tool for ICU rounding teams was developed by King et al \cite{king2023voice}, that utilized real-time discussions to prompt evidence-based practices. FM-based chatbots in languages other than English have also been adopted \cite{blanc2022flaubert, wael2021intelligent}. Incorporating FM-based medical chatbots and virtual assistants can improve healthcare accessibility and help relieve the burden on the healthcare system by facilitating communication between patients and healthcare providers. For instance, providing scalable and personalized patient support, augmenting clinical decision-making processes, and providing expert input to general practitioners when specialists are not available.

\subsubsection{Content generation (clinical reports)}
FMs have shown promising performance and assistance capability, especially in generating radiology reports. For instance, Clinical-BERT \cite{clinicalbert} can accurately predict disease from radiographs and generate reports.
Similarly, RadioBERT \cite{radiobert}, proposed by Kaur and Mittal, generates accurate medical reports from CXRs. Beyond BERT-based approaches, including \cite{medicalVlBert}, GPT-based models have also achieved promising performance. For instance, Nakaura et al. \cite{nakaura2023preliminary} evaluated the performance of GPT models in generating radiology reports based on patient information and imaging findings, and their performance was compared with reports generated by radiologists. Luo et al. \cite{biogpt} proposed a GPT-s-based model, BioGPT, that is specifically fine-tuned for various biomedical tasks, including end-to-end RE and QA. 
These advancements demonstrate the potential of FMs to improve diagnostic accuracy and assist radiologists in automating the reporting processes.

\subsubsection{Information extraction}
Xu et al. \cite{xu2020unified} proposed a BERT-based approach to link text phrases to ontology concepts, which can be employed in various downstream tasks including RE and information retrieval. The proposed BERT-based approach achieved an accuracy of 83.5\% in linking phrases to Unified Medical Language System (UMLS) concepts. Pan et al. \cite{pan2021bert} proposed a BERT-based medical text–to-SQL model (MedTS) to translate medical texts into SQL queries for Electronic Medical Records (EMRs) that achieved promising performance compared to existing techniques. Hanna et al. \cite{hanna2023assessing} utilized ChatGPT to racial bias. The authors used ChatGPT to generate healthcare text for HIV patients, analyzing 100 electronic health record encounters with varied races/ethnicities and finding no significant differences in sentiment or subjectivity. A BERT-based model was proposed by Kuling et al. \cite{kuling2022bi} to segment breast radiology reports into different sections 
that achieved 98\% accuracy. 
FMs could be beneficial in extracting useful information from unstructured medical text data. Furthermore, such robust models offer promising avenues for improving data interoperability and accessibility in healthcare settings.

\subsubsection{Information retrieval (bedside decision support)}

Clinical decision support (CDS) plays a vital role in delivering information and tailored recommendations to patients and healthcare providers during care delivery. FMs can be adopted to significantly improve healthcare delivery. For instance, Liu et al. \cite{liu2023using} showed that ChatGPT could be a valuable complement in optimizing CDS alerts and advancing learning health systems. FM applications can facilitate accurate information retrieval to provide the most comprehensive and pertinent historical information to the provider. 

\subsubsection{Clinical Language translation}
Medical language translation aims to transform complex medical texts such as those embedded in EHRs into a clearer format for diverse audiences \cite{li2022neural}. This is achieved by simplifying complex medical terminology, abbreviations, and technical language so that they can be more easily understood by patients, caregivers, and individuals lacking specialized medical knowledge. For instance, ``peripheral edema'' or ``hypertension'' may be replaced with or tagged as ``ankle swelling'' or ``high blood pressure'', respectively \cite{li2022neural}. FMs show high accuracy for contextual medical language translation \cite{li2022neural}. By accurately contextualizing medical terminology and simplifying complex language, FMs enable clearer and more accessible communication between healthcare providers, patients, and caregivers. 

\subsection{Medical Computer Vision}
\label{computervision}

\subsubsection{Generation}
FMs have been widely adopted in the computer vision domain for multiple tasks, including content generation. For instance, Khader et al. \cite{medicaldiffusion} implemented diffusion probabilistic models to synthesize high-quality 3D MRI and CT images. The results were validated by radiologists, proving that the model generates 3D volumetric medical imaging data that accurately reflects human anatomy. A stable diffusion-based model, RoentGen, for synthesizing chest X-rays (CXR) conditioned on text prompts was proposed by Delbrouck et al. \cite{roentgen}. The proposed RoentGen model can generate high-fidelity synthetic medical images with controllable features. Another work utilizing stable diffusion was carried out by Kidder \cite{kidder2023advanced}, in which photorealistic medical images were generated including MRI, chest, and lung X-rays, and contrast-enhanced spectral mammography (CESM) images.
Similarly, Moon et al. \cite{moon2022multi} proposed the Medical Vision Language Learner (MedViLL), which combines textual representations (BERT) and visual representations for multiple downstream tasks such as diagnosis classification, medical image-report retrieval, and radiology report generation. Yang et al. \cite{yang2023customizing} used the ImageCLEF 2023 dataset for medical report generation using BLIP-2 \cite{blip2} model.
FMs could generate realistic data for training and validating algorithms, thus reducing the use of expensive and sensitive real-world datasets. Furthermore, synthetic data can support the development of cutting-edge models without putting any patient information at risk. 

\subsubsection{Classification}
DL models enable the prediction of predefined labels or categories associated with input data. The results are promising and often outperform medical experts in terms of accuracy \cite{tiu2022expert}. Since DL-based models rely heavily on annotated datasets, they are not always suitable for generalization. Hence, FMs that utilize SSL or zero-shot classification can be a promising alternative. For instance, Tiu et al. \cite{tiu2022expert} proposed CheXzero which demonstrated the effectiveness of SSL for pathology classification on CXR images without explicit annotations, achieving comparable accuracy to radiologists. Similarly, attention-based efficient Global-Local Representations for Images using the Attention (Gloria) method \cite{gloria} contrasts image sub-regions and words in paired reports to global and local multimodal representations. Wang et al. \cite{wang2022medclip} showed that decoupling image-text pairs effectively scale the training data and reduces the rate of false positives. However, models such as CheXzero and MedCLIP treat partial positive pairs as entirely negative pairs. Although MedCLIP reduces this issue, it still relies on expert labels for high performance. Therefore, Jang et al. \cite{jang2022significantly} proposed a fine-tuning strategy by producing sentence sampling and positive-pair loss relaxation to achieve significant improvements (over 5\%) in zero-shot pathology classification across various CXR datasets. You et al. \cite{cxrClip} proposed CXR-CLIP, which utilizes image-text pairs with general prompts and multiple images and report sections. 
A diffusion-based model, DiffMIC, was proposed by Yang et al. \cite{yang2023diffmic} to eliminate perturbations and noise  in medical images and robustly capture semantic representations.
The accuracy generalization of the classification models across various domains can be significantly improved by integrating techniques such as SSL, and attention-based frameworks. Furthermore, the development of diffusion-based models highlights the importance of robust semantic representation in medical imaging.

\subsubsection{Segmentation}
\label{segmentation}
Medical image segmentation plays a vital role in medical diagnostics and provides sufficient non-invasive information to visualize, analyze, and track anatomical structures for accurate diagnosis and treatment planning \cite{elnakib2011medical}. 
Recent ViT-based models include Swin UNEt TRansformers (Swin UNETR), proposed by Hatamizadeh et al. \cite{Swinunetr}, which is a novel semantic segmentation model for brain tumor images that uses a sequence-to-sequence prediction approach. Swin UNETR combines a U-shaped network architecture with a Swin Transformer and has achieved SOTA performance in the BraTS 2021 segmentation challenge. 
Cox et al. \cite{cox2024brainsegfounder} proposed SwinUNETR-based BrainSegFounder, which is a 3D medical FM for neuroimage segmentation. The authors show that the proposed BrainSegFounder reduces the need for labeled data and achieves high segmentation accuracy, outperforming prior models on benchmark datasets.

Butoi et al. \cite{universeg} showed that segmentation can be challenging due to the limited training data and the complex structures of human organs. Furthermore, DL-based models face challenges in generalization \cite{samclip}. Therefore, they proposed a universal medical image segmentation (UniverSeg) model trained on a large dataset of over 22,000 scans collected from 53 open-access datasets from 16 different modalities. The proposed UniverSeg achieved promising performance compared to SOTA models such as SENet \cite{SENet}, PANet \cite{panet}, and ALPNet \cite{ALPNet}. Kim et al. \cite{kim2023empirical} showed that UniverSeg can be generalized for various tasks with minimal labeled data. Similarly, an application of SAM was evaluated for zero-shot tumor segmentation, tissue segmentation, and cell nuclei segmentation \cite{deng2023segment}. The performance of the SAM model was better for large connected objects but faced challenges in dense instance object segmentation. Another SAM-based model was also used for interactive medical image segmentation \cite{shen2023temporally}.

Xing et al. \cite{DiffUNet} showed that integrating denoising diffusion models (DDMs) into a U-shaped network architecture can improve the volumetric segmentation for multiple modalities. A DDM-based model was also proposed for dental radiography segmentation \cite{rousseau2023pre}. Another denoising diffusion probabilistic model (DDPM)-based \cite{ho2020denoising} model that incorporated contextual discounting for tumor segmentation was used for multi-modality medical images. An improved diffusion model was proposed by Zhao et al. \cite{dtan} that combines text attention principles with diffusion models for enhanced medical image segmentation precision.
Wolleb et al. \cite{wolleb2022diffusion} introduced a novel semantic segmentation method for medical images that generates image-specific segmentation masks and computes pixel-wise uncertainty maps. The authors utilized diffusion models \cite{nichol2021improved} using the BRATS2020 dataset to achieve meaningful segmentation masks that improved brain tumor segmentation performance compared to SOTA models. Rahman et al. \cite{rahman2023ambiguous} proposed a diffusion-based segmentation approach to synthesize multiple variants of segmented masks based on various radiologist annotations. The authors proposed the Ambiguity Modelling Network (AMN) and Ambiguity Controlling Network (ACN) to capture uncertainty in ground truth and predicted masks. Furthermore, the Collective Insight (CI) score, combining sensitivity, the maximum Dice matching score, and the diversity agreement score was proposed. MedSegDiff, based on diffusion probabilistic models (DPMs), was proposed by Wu et al. \cite{medsegdiff} for segmentation from MRI and ultrasound images. An extended version of MedSegDiff-V2 was proposed by Wu et al \cite{medsegdiff2}. MedSegDiff-V2 integrated ViT into DPM and was evaluated for 20 segmentation tasks. Another diffusion model for vessel segmentation was proposed by Kim et al. \cite{kim2022diffusion} by combining the DDPM \cite{ho2020denoising} with adversarial learning to diagnose vascular diseases. 
The results evaluated on multiple datasets achieved generalized and superior segmentation performance. Fu et al. \cite{fu2023importance, fu2023recycling} presented denoising diffusion models that predict segmentation masks using a fully noise-corrupted input rather than relying on conventional ground truth masks during training. Sun et al. \cite{sun2023enhancing} proposed a diffusion-model-based method to automatically segment gland instances in histological colon images.
We outlined CLIP-based applications in Section~\ref{CLIP}, but some CLIP-driven applications specifically for image segmentation include those of \cite{poudel2023exploring}. These models showed that CLIP-based approaches exhibit reasonable transferability to medical images, especially in zero-shot settings for non-radiology photographic modalities. Similarly, Adhikari et al. \cite{adhikari2023synthetic} demonstrated the performance of CLIPseg and CRIS on synthetic data obtained from diffusion models. 
FMs can assist by enhancing diagnostic accuracy, surgical planning, disease staging, and medical decision-making for risk stratification. It can help to achieve faster diagnoses, tailored treatment decisions, and improved patient outcomes. Furthermore, incorporating FMs in clinical settings can assist clinicians in further refining segmentation accuracy and efficiency to improve the precision in various medical imaging applications.

\subsubsection{Localization}
FMs can serve as digital assistants to radiologists by drafting reports from imaging modalities which will substantially reduce their workloads \cite{moor2023foundation}. Such automatically generated detailed reports can be used to identify abnormalities and contextualize findings based on the patient's medical history \cite{moor2023foundation}. Furthermore, FM-based grounded radiology reports can assist radiologists in providing more informed interactive conversations. Such as a radiologist might ask the model to identify new multiple sclerosis lesions compared to previous images \cite{moor2023foundation}. Moreover, region-guided radiology report generation models enhance report completeness and consistency by focusing on individual anatomical regions and generating tailored descriptions \cite{tanida2023interactive}. 
Zhu et al. \cite{zhu2023visual} proposed a novel visual grounding framework for multimodal fusion by aligning image-text embedding in a shared semantic space and utilizing a transformer-based deep interactor for accurate reasoning.
However, implementation of such generalized models on a large scale may require the integration of data from multiple sources such as patient history, laboratory results, and previous imaging studies to provide contextually relevant insights \cite{moor2023foundation}.
Visual grounding from imaging modalities saves time and ensures consistency and completeness in reporting. Moreover, by contextualizing findings based on patient history, FMs facilitate more informed discussions between radiologists and patients.

\subsubsection{Reconstruction}
High-quality images are essential for clinical diagnosis, but addressing concerns related to cost and patient safety is also important \cite{shokrollahi2023comprehensive}. Various imaging modalities, such as CT, MRI, and PET, require advanced reconstruction techniques to enhance spatial resolution, reduce noise, and improve overall image quality \cite{guo2023reconformer}. FMs have been shown to be effective for image reconstruction tasks. For example, Levac et al. \cite{levac2022accelerated} proposed deep generative diffusion models to address inverse problems in accelerated image reconstruction with subject motion during MRI examinations. Similarly, Gungor proposed AdaDiff \cite{adaptivediffusion} for adaptive diffusion before MRI reconstruction. AdaDiff utilizes a diffusion process with an adversarial mapper for training and employs a two-phase reconstruction approach. The proposed approach is resilient against domain shifts in multi-contrast brain MRIs and has achieved promising performance for accelerated MRI reconstruction. The results from other works show that diffusion modeling is a promising approach for MRI reconstruction \cite{korkmaz2023self, kazerouni2023diffusion}. FMs can improve healthcare accessibility by augmenting or enhancing image quality, particularly in facilities lacking advanced hardware or scanning capabilities and technical expertise.

\subsubsection{Enhancement}
Image enhancement methods using super-resolution techniques can enhance the quality and clarity of medical images. Super-resolution methods aim to enhance fine details by reconstructing high-resolution images from lower-resolution inputs. In medical imaging, they aim to provide precise visual information to clinicians \cite{umirzakova2023medical}. Recently, multiple studies have been carried out on super-resolution techniques \cite{umirzakova2023medical}. For instance, Mao et al. \cite{mao2023disc} proposed the Disentangled Conditional Diffusion Model (DisC-Diff) that incorporates a disentangled U-Net, forward and reverse diffusion processes, and novel loss functions to enhance the model's effectiveness in capturing complex interactions in multi-contrast MRI. The proposed approach achieved promising performance compared to other works \cite{dhariwal2021diffusion, liang2021swinir}. Another robust latent diffusion-based model proposed in \cite{wang2023inversesr} showed promising performance in clinical MRI scans characterized by low resolution, diverse contrast, and spatial resolution variations.
FMs significantly improve the quality and clarity of diagnostic images, enabling clinicians to make more informed decisions that lead to better patient outcomes. For instance, FMs could allow hospitals with only 1.5T MRIs to synthesize 7T data, which would greatly affect clinical care and diagnostic accuracy.

\subsubsection{Modality Translation}
Multi-modality imaging such as CT-MRI provides key information for an improved decision-making process. However, obtaining multi-modality images can be expensive and requires acquisition time \cite{yan2022swin}. Therefore, synthesizing missing or corrupted modalities from available ones can provide a cost-effective medical image-to-image translation approach. Many DL-based models have been employed for this purpose, particularly GANs. 
However, it has recently been shown that diffusion models are promising alternatives to GANs providing better results \cite{ozbey2023unsupervised, jiang2023cola}. For instance, SynDiff \cite{ozbey2023unsupervised} employs a conditional diffusion process that effectively translates between source and target modalities by progressively mapping noise and source images onto the target image. Another diffusion-model-based approach utilizes 2D backbones for volumetric data synthesis \cite{zhu2023make}.
FMs can be a promising approach to synthesizing multi-modality imaging and being able to translate information from one modality to another. This will assist clinicians with enhanced diagnostic capabilities and facilitate more informed treatment decisions.

\subsubsection{Augmented procedures}
Incorporating augmented reality (AR) into medical interventions and processes empowers healthcare providers with real-time, contextually relevant insights to improve decision-making and overall patient care. 
Augmented procedures can be applied across a spectrum of medical scenarios, including neurosurgery training, robotic neurosurgery, pain management, neuronavigation, rehabilitation, consent taking, and diagnostic tools \cite{ghaednia2021augmented}. By incorporating multi-modality data, FMs can potentially provide real-time assistance to surgical teams by assisting in various tasks including annotating video streams, alerting to skipped procedural steps, and diagnosing rare anatomical phenomena.

\subsubsection{Vision-based conversation}
Conversational agents (chatbots) gained significant attention, especially after ChatGPT in various domains including healthcare. Traditional chatbots in healthcare primarily relied on textual inputs from patients to aid and information. However, incorporating vision data can enhance the capabilities of these conversational agents, enabling them to understand and respond to visual cues effectively. For instance, Luo et al. \cite{luo2023valley} proposed a multi-model FM, Valley (Video Assistant with Large Language model Enhanced ability), capable of understanding video, image, and language simultaneously. Qilin-Med-VL 
\cite{liu2023qilin} is a Chinese large vision-language model for healthcare to comprehend both textual and visual medical data. Similarly, healthcare-based smart chatbots \cite{pires2024robodoc} intelligently respond to user input to detect conditions like viral and bacterial pneumonia in X-ray images and Drusen, Choroidal Neovascularization (CNV), and Diabetic Macular Edema (DME) in Optical Coherence Tomography (OCT) images. Furthermore, models such as VideoChat \cite{li2023videochat}, VAST \cite{chen2023vast}, visualGPT \cite{wu2023visual}, NExT-GPT \cite{wu2023next}, CLIPSyntel \cite{ghosh2023clipsyntel}  can aid in analyzing images or videos shared by patients to gather valuable insights into their physical condition, environment, symptoms to offer more personalized and context-aware support.
FMs offer valuable assistance in analyzing visual information shared by patients to provide enhanced communication and diagnostic capabilities in healthcare settings.

\subsection{Graph Learning}
\label{graphLearning}
LLMs have excelled in natural language tasks but face challenges in graph learning due to their limitations in mathematical precision, logic reasoning, spatial perception, and the handling of temporal information. Therefore, work has been done to develop a framework specifically for graph data. Significant approaches include incorporating the capabilities of ChatGPT and the Toolformer \cite{zhang2023graph} model for efficient graph reasoning \cite{zhang2023graph}. Furthermore, incorporating multi-modality data provides better performance in graph learning models \cite{liu2023git}. 

\subsubsection{Brain network}
Brain networks are complex and interconnected neural connections within the human brain \cite{murugesan2020brainnet}. Modern imaging technologies such as functional MRI (fMRI) have provided insight into normal and abnormal brain functions \cite{murugesan2020brainnet}. Graph neural networks (GNNs) show great potential in capturing complex interpretable brain network models for disease prediction \cite{cui2022braingb}. Furthermore, GNN models are more common for investigating the brain connectome and how disruptions or alterations in brain networks are associated with conditions such as Alzheimer's disease \cite{zhou2022interpretable}, Parkinson's disease \cite{huang2023mnc}, psychiatric disorders \cite{zheng2024ci}, rare diseases \cite{sanjak2024clustering}, and autism spectrum disorders \cite{menon2023asdexplainer}.
Due to their generalized nature, FMs offer a promising alternative to traditional GNNs for extracting meaningful patterns from brain data. FMs' ability to capture complex interactions within brain networks makes them well-suited for deciphering neural connections associated with neurological disorders.

\subsubsection{Knowledge discovery in medicine}
Knowledge discovery in medicine, also termed heuristics refers to the process of extracting valuable insights, patterns, and knowledge from large and complex medical datasets \cite{hasselgren2024artificial}. FMs are also adopted for knowledge discovery and drug discovery with great success \cite{edwards2023synergpt}.
Utilizing FMs in medical knowledge discovery can reveal novel correlations, biomarkers, and therapeutic targets for drug development, personalized medicine, and healthcare delivery.

\subsection{Biology and Omics}
\label{bioandomics}

FMs focused on biology and omics were outlined in Section~\ref{omics}. However, this section briefly explains the details of their applications. Additionally, we refer readers to several articles for detailed works concerned with the biology and omics domain: \cite{madani2023large,lin2023evolutionary,ferruz2022protgpt2,chowdhury2022single, theodoris2023transfer}.

\subsubsection{Drug discovery}
Drug discovery aims to identify and develop new therapeutic compounds for treating various medical conditions. Traditional drug discovery methods have limitations due to small datasets and generalization challenges. However, the authors in\cite{cao2023instructmol,mendez2022mole} demonstrated the potential of generative drug design and molecular FMs. They achieved improved molecular quality and an enhanced capability for predicting drug properties. FMs could decrease the cost, time, manpower, and hardware requirements compared to traditional drug discovery and development. Furthermore, it also can impact high-throughput workflows.

\subsubsection{Radiomics}
Radiomics mainly involves extracting and analyzing quantitative features from medical images to provide valuable insights into lesion characterization, prognosis, and treatment response \cite{mayerhoefer2020introduction}. Radiomics enables a more comprehensive and objective assessment of the underlying tissue properties by transforming conventional medical images (CT, MRI, PET) into high-dimensional data sets \cite{gillies2016radiomics}. A work done by Pai et al. \cite{pai2023foundation} proposed a SimCLR-based model \cite{chen2020simple} for managing biomarker discovery through self-supervised learning on a dataset comprising 11,467 radiographic lesions. The results show that FMs are efficient in learning imaging biomarkers and are accelerating their translation into clinical settings.
FMs demonstrate remarkable potential in identifying imaging biomarkers and facilitating their translation into clinical practice. These capabilities have significant clinical implications such as better disease characterization, prognosis assessment, and personalized treatment strategies.

\subsubsection{Proteomics}
In 1995, Marc Wilkins created the term ``proteome'' for a new type of omics \cite{al2021proteomics}. Proteomics investigates the interactions, compositions, functions, and structures of proteins and their cellular activity \cite{al2021proteomics, wilkins1996progress}. Many works have been carried out to develop intelligent solutions for proteomics\cite{al2021proteomics, wilkins1996progress}. Some of the relevant FMs include \cite{ferruz2022protgpt2,chowdhury2022single, rives2021biological, lin2023evolutionary}.
FMs can unlock new insights into complex cellular activities, paving the way for innovative approaches in drug discovery, personalized medicine, and disease management.

\subsection{Other Applications}
\label{others}

\subsubsection{Text-to-Protein generation}
Protein design aims to create customized proteins for various applications to address environmental and biomedical challenges. Leveraging recent advancements in transformer-based architectures has resulted in promising performance. For instance, leveraging contrastive learning and generative models, Liu et al. \cite{liu2023text} proposed ProteinDT for text-guided protein generation to property prediction. Trained on a large-scale dataset (SwissProtCLAP, with 441K text and protein pairs), it outperformed SOTA methods in protein property prediction benchmarks \cite{rao2019evaluating, rao2021msa, zhang2022ontoprotein}. ProtGPT2 \cite{ferruz2022protgpt2} was introduced as a language model that generates novel protein sequences following natural principles that exhibit amino acid propensities similar to natural proteins. ProGen \cite{madani2020progen} generates protein sequences exhibiting predictable functions within broad protein families. Similarly, generalist medical AI (GMAI) \cite{moor2023foundation} is a model capable of generating both amino acid sequences and three-dimensional structures of proteins based on textual prompts. Watson et al. \cite{watson2022broadly} proposed a diffusion-based model for protein backbone design that fine-tunes the RoseTTAFold structure prediction network on protein structure denoising tasks.
FMs present an innovative approach to generating protein sequences and structures from textual prompts. By integrating in-context learning capabilities, FMs could dynamically define tasks based on example instructions, such as generating proteins with specific binding properties.

\subsubsection{Video}
FMs were also successful in extracting insightful information from videos \cite{arnab2021vivit,liu2022video,luo2023valley}. For instance, Wang et al. \cite{endoFM} proposed the Endo-FM model for endoscopic video analysis based on a video transformer. The work used an SSL pre-training approach that captures spatial and temporal dependencies, pre-trained in a self-supervised manner on a large-scale dataset of 5 million frames obtained from 33K videos. A CLIP-based model, EchoCLIP \cite{christensen2023multimodal} for echocardiography trained on a dataset of over a million cardiac ultrasound videos and expert interpretations demonstrated robust zero-shot performance in cardiac function assessment and identification of intracardiac devices. FMs enable comprehensive understanding and analysis of dynamic visual data to enhance decision-making and diagnostic accuracy in healthcare.

\subsubsection{Speech}
Multiple works are done to utilize FMs for speech audio information. For instance, AudioGPT is based on the GPT model that integrates audio FMs and input/output interfaces for spoken dialogue \cite{huang2023audiogpt}. Similarly, various works show that FMs can complement the processing of speech data \cite{hsu2021hubert, rubenstein2023audiopalm}. For instance, a study done by Gerczuk et al. \cite{gerczuk2023zero} detects smaller temporal variations in depressed mood by introducing a zero-shot personalization strategy for large speech FMs.
FMs offer valuable capabilities for analyzing speech data and extracting meaningful insights that are not readily perceivable by healthcare providers.

\onecolumn
\begin{center}
\begingroup
\scriptsize 
\begin{longtable}{p{0.6cm} p{0.6cm} p{1.3cm} p{1.5cm} p{1cm} p{3.9cm} p{1.5cm} p{0.6cm} p{3cm}}
    \caption{Foundation Models in Healthcare domain, presented in chronological order.}
    \label{tab:Flagship} \\
    \toprule
     \scriptsize
    \textbf{Work} & \textbf{Year} & \textbf{Model} & \textbf{Base Model}& \textbf{Parameter}  & \textbf{Dataset} & \textbf{Dataset Size }&\textbf{Public} & \textbf{Task} \\
    \midrule
    \endfirsthead
    \caption[]{Foundation Models in Healthcare domain (continued)} \\
    \toprule
    \textbf{Work} & \textbf{Year} & \textbf{Model}& \textbf{Base Model} & \textbf{Parameter}  & \textbf{Dataset}  & \textbf{Dataset Size }&\textbf{Public} & \textbf{Task} \\
    \midrule
    \endhead
    \bottomrule
    \endfoot 
    \bottomrule
    \endlastfoot

\cite{biobert} & 2018 & BioBERT 
   & BERT
   &345M & English Wikipedia, BooksCorpus, PubMed Abstracts, PMC Full-text articles & 4.5B words 
   & \href{https://github.com/dmis-lab/biobert} {\cmark} 
   & Biomedical Text Mining   \\

\cite{biomegatron} & 2020 &  BioMegatron 
     &  MegatronLM \cite{shoeybi2019megatron}
     &1.2B  & PubMed abstract, CC0-licensed Commercial Use Collection of the PMC full-text corpus & 6.1B words  & \href{https://github.com/NVIDIA/NeMo}{\cmark} & NER, QA, RE  \\

\cite{rnabert} & 2022  &  RNABERT 
    & BERT
    & - & SRNAcentral (human-derived small ncRNAs) & 76237 ncRNAs &
    \href{https://github.com/mana438/RNABERT} {\cmark} & ncRNA structural alignment and clustering.  \\

\cite{scbert} & 2022  &  scBERT 
     & BERT
     & - & Seven publicly available scRNA-seq datasets & $>$ 1.8M cells & 
      \href{https://github.com/TencentAILabHealthcare/scBERT} {\cmark} & Cell type annotation.  \\  

\cite{rnaFM} & 2022  &  RNA-FM 
    & BERT
    & & RNAcentral \cite{rnacentral2021} & $>$ 23M sequences& 
    \href{https://github.com/ml4bio/RNA-FM} {\cmark} & RNA biology.  \\

\cite{GatorTron} & 2022	& GatorTron	
        & Megatron BERT
        & 8.9B 	& UF Clinical notes, PubMed articles, Wikipedia& $>$90B words & \href{https://catalog.ngc.nvidia.com/orgs/nvidia/teams/clara/models/gatortron_og} {\cmark}	&NER, QA, RE, NLI, semantic textual similarity \\

\cite{medpalm} & 2023 & Med-PaLM 
        & Flan-PaLM \cite{chung2024scaling}
        & 540B & MultiMedQA benchmark & 780B tokens & \xmark & Medical QA  \\
        
\cite{spliceBERT} & 2023  &  SpliceBERT 
    & BERT
    & 19.4M & 2 million pre-mRNA
sequences & 2M & 
\href{https://github.com/chenkenbio/SpliceBERT} {\cmark} & RNA splicing prediction.  \\

\cite{rnaMSM} & 2023  &  RNA-MSM 
     & BERT
     & - & rFam \footnote{\url{https://rfam.org/}} & 4069 RNA families &
     \href{https://zenodo.org/records/8280831} {\cmark} & multiple sequence alignment.  \\

\cite{multitalent} & 2023 & MultiTalent 
        & SwinUNETR,  3D U-Net
        &  69M & 13 public abdominal CT datasets &1477 3D images  &  
        \href{https://github.com/MIC-DKFZ/MultiTalent} {\cmark} &  image segmentation\\

\cite{scgpt} & 2023  &  scGPT 
    & GPT
    & 53M & 8 datasets     & $>$ 10M cells&
    \href{https://github.com/bowang-lab/scGPT} {\cmark} & Single-cell biology.  \\

\cite{medpalm2} & 2023 &  Med-PaLM2  
        & PaLM 2 
        & - & MultiMedQA benchmark   & - & \xmark & Medical QA  \\
        
 \cite{GatorTrongpt} & 2023 &  GatorTronGPT 
        & GPT-3, MegaTronLM 
        &  20B  & UF Clinical notes, Pile dataset  & 277B words& \href{https://www.nature.com/articles/s41746-023-00958-w#code-availability} {\cmark} & NER, QA, RE, NLI, semantic textual similarity \\
        
\cite{Sammed3d}  & 2023 &  SAM-Med3D 
      &
      & 100M &21K medical images & 121K masks &   \href{https://github.com/uni-medical/SAM-Med3D}{\cmark} & Medical 3D image segmentation  \\

 \cite{endoFM} & 2023 &  Endo-FM 
        &Video transformer
        & 121M & 7 datasets for pre-training  & 5 million frames&
        \href{https://github.com/med-air/Endo-FM} {\cmark} & Endoscopy Video Analysis  \\

\cite{getModel} & 2023  &  GET 
    & Scratch
    & 80M & Single-cell Chromatin accessibility data &213 cell types &
    \href{https://github.com/GET-Foundation} {\cmark} & Transcription regulation prediction.  \\
         
 \cite{bigRNA} & 2023  &  BigRNA 
    &
    & 
    & GTEx consortium &5,201 RNA-Sequence &\xmark & RNA biology.  \\

 \cite{hosseinzadeh2023towards} & 2023 &  Adam 
       & Scratch
       & 25M &  CXR-14 \cite{wang2017chestx} and EyePACS \cite{cuadros2009eyepacs} & $>$108K images &
       \href{https://github.com/MR-HosseinzadehTaher/Eden} {\cmark} & Human organ identification.  \\
       
 \cite{genecompass} & 2023  &  GeneCompass 
     & Scratch
     & 100M & scCompass-126M (transcriptomes from human and mice) & 120 M single-cell transcriptomes & 
     \href{https://github.com/xCompass-AI/GeneCompass} {\cmark} & Cross-species genomics.  \\
     
\cite{schyena} & 2023  &  scHyena 
    &
    &
    &GTEx consortium & &\xmark \footnote{Link not working} & scRNA-seq analysis in the brain.  \\

\cite{meditron} & 2023 & MEDITRON 
        & Llama-2
        & 70B  &Clinical Guidelines, PubMed, Experience Replay & 48.1B tokens& 
        \href{https://github.com/epfLLM/meditron} {\cmark} & \\

\cite{du2023segvol} & 2023 &  SegVol 
        & Scratch
        & 180M&  25 CT datasets with 5772 CTs & 90K unlabeled CTs & 
        \href{https://github.com/BAAI-DCAI/SegVol} {\cmark} &  Image segmentation \\

\cite{genept} & 2023  &  GeneGPT 
    &ChatGPT
    &  
    & 33000 genes & & 
    \href{https://github.com/yiqunchen/GenePT} {\cmark} & Genomics and single-cell transcriptomics.  \\
    
\cite{dnabert2} & 2023  &  DNABERT2 
    & BERT
    & 117M &32.49 billion nucleotide from 135 species & 32.49B nucleotide bases &
    \href{https://github.com/MAGICS-LAB/DNABERT_2} {\cmark} & Genome FM.  \\
    
\cite{GexMolGen} & 2023  &  GexMolGen 
    & scGPT
    & 
    & L1000 public dataset & &
  \href{https://github.com/Bunnybeibei/GexMolGen} {\cmark} & Hit-like molecules, drug discovery.  \\

\cite{gfetm} & 2023  &  GFETM 
    & DNABERT
    & 2.5B &GTEx consortium &$>$ 10K cells &\
    \href{https://github.com/fym0503/GFETM} {\cmark} & scATAC-seq Analysis.  \\

\end{longtable}
\endgroup
\end{center}

\begingroup
\scriptsize 
\begin{longtable}{p{0.7cm} p{2.9cm} p{2.1cm} p{3cm} p{1cm} p{2cm} p{3.8cm}} 
    \caption{Applications of FMs in healthcare. CNLP: Clinical NLP, MCV: Medical computer vision. The datasets can be accessed by clicking the embedded link \cmark of each column. Multiple \cmark for each work indicates links to multiple datasets.} \\
    \toprule
    \textbf{Work} &  \textbf{Task} & \textbf{Modality} & \textbf{Dataset} & \textbf{Link to Dataset}& \textbf{Pretrained Model} & \textbf{Keypoints} \\
    \midrule
    \endfirsthead 
    \caption[]{Applications of FMs in healthcare (continued)} \\
    \toprule
     \textbf{Work} & \textbf{Task} & \textbf{Modality} & \textbf{Dataset} & \textbf{Link to Dataset}& \textbf{Pretrained Model} & \textbf{Keypoints} \\
    \midrule
    \endhead 
    \bottomrule
    \endfoot 
    \bottomrule
    \endlastfoot 

\cite{hamidi2023evaluation} & 	Evluation of ChatGPT and Claude&	EHR&	MIMIC III & \href{https://www.ncbi.nlm.nih.gov/pmc/tools/openftlist/} {\cmark} &	ChatGPT and Claude	& Accurate response \\

\cite{shahsavar2023user} &		ChatGPT for self-diagnosis&	Questionnaire&	Questionnaire
&	\xmark
& ChatGPT	& Majority participants were satisfied from ChatGPT \\

\cite{bernstein2023comparison} &  Chatbots evaluation against ophthalmologist&	Question-answer pairs	&Eye Care Forum (question-answer pairs)	
& \href{https://cdn.jamanetwork.com/ama/content_public/journal/jamanetworkopen/939195/zoi230872supp2_prod_1691786958.13885.pdf?Expires=1730019784&Signature=DfMUmU-x3AN6R3tFEWCusRyzBg1lUp9exc4Og3EX4KOwJghrD0UPV7F15IkFiqSvsYQQIIMYMiSkSXDoQugHv591aKeJKKjOIT-tBzIsucgKW5uCMTbOyKUorBRhM-Zc7f2UsyFL-ZmAKWkckHcIdjksiJzwE9lTMzJ6VTKEZsMICy5NrfFt6yR7XR-MPEGrgyeCG~v2rPm3qdMWz588NpxR2b2TfgbFjAuQIr9nw8qXAt5zt2D6J4gNT8NY-C0W712eVNCkbkOgqpDB9iMvMKWWx8vVrZ~UcjJCxpSsydPqXkJPD3aCtDXvtYMxiFc6sIBTZtAX6EqELlsJgOn7mA__&Key-Pair-Id=APKAIE5G5CRDK6RD3PGA} {\cmark}
& ChatGPT	& ChatGPT provide comparabable quality advice\\

\cite{huang2023assessment} &	Chatbots (chatGPT 3.5, 4) evaluation against Family Medicine residents.&	Multiple-choice questions (MCQs)&	MCQs
&	\xmark
& ChatGPT (3.5 and 4)&	GPT-4 performed the best.\\

\cite{sun2023pathasst} &	Diagnostic \& Predictive Analytics &	Multimodal (image-text pairs)&	PathCap and PathInstruct \cite{sun2023pathasst}  
&	\href{https://huggingface.co/datasets/jamessyx/PathCap} {\cmark}
\href{https://huggingface.co/datasets/jamessyx/PathInstruct} {\cmark}
&Vicuna-13B \cite{chiang2023vicuna}, CLIP (PLIP) \cite{huang2023leveraging} & Improved pathology diagnosis and treatment processes\\

\cite{king2023voice} &Digital assistant evaluation. &	Audio signals&	106 patient audio recordings, PubMed Central manuscripts, and MIMIC III & \xmark 
& BioClinicalBERT	& FMs aid in reduced prompts per patients. \\

\cite{xu2020unified} & Concept normalization&	Text	& 1000 discharge summaries \cite{uzuner20112010}
& \href{https://i2b2.org/NLP/DataSets} {\cmark}
&BERT	&Improved Accuracy \\
\cite{pan2021bert} &	Medical text to SQL queries	&EHR&	MIMICSQL & 
 \href{https://github.com/wangpinggl/TREQS} {\cmark}
&BERT& Improved text-to-SQL generation
\\

\cite{medicaldiffusion} &	3D MRI and CT&	Images	& MRNet Dataset, ADNI, Breast Cancer MRI Dataset, LIDC, IDRI, 
&	\href{https://www.nature.com/articles/s41598-023-34341-2#data-availability} {\cmark}
&VQ-GAN DDPM  & High quality 3D volumetric data synthesis \\

\cite{roentgen} &	Chest X-ray (CXR) synthesis &	Multimodal (image-text pairs)&	MIMIC-CXR dataset 
& \href{https://physionet.org/content/mimic-cxr/2.1.0/} {\cmark}
&Stable Diffusion	& High-fidelity synthetic  CXRs conditioned on text.\\

\cite{moon2022multi} &	Vision-Language Tasks	& Multimodal (image-text)&	MIMIC-CXR,VQA-RAD, and Open-I 
& \href{https://physionet.org/content/mimic-cxr/2.1.0/} {\cmark} 
 \href{(http://openi.nlm.nih.gov/} {\cmark} 

&BERT, ResNet-50	 & Multi-modal understading of text-image pairs \\

\cite{yang2023customizing}	& Report generation&	Multimodal (image-text pairs)	& ImageCLEF 2023	
&  \href{https://github.com/razorx89/roco-dataset} {\cmark} 
&BLIP-2, EVA-ViT-g as encoder, ChatGLM-6B as decoder	&FMs improved performance. \\

\cite{kidder2023advanced} &	MRI, CXRs, CESM	&Multimodal (image-text pairs)&	\cite{khaled2022categorized,buda2019association,cheng2016retrieval,cheng2015enhanced} 
&\href{https://academic.oup.com/biomethods/article/9/1/bpae062/7739892#480792604} {\cmark} 
&Stable Diffusion	& Generate photorealistic medical images\\

\cite{hanna2023assessing}	&	Evaluate FMs for racial bias 	&Text	&100 EHR encounters
&	 \xmark
&ChatGPT&	No racial bias \\

\cite{kuling2022bi}	& Section segmentation &	Text (radiology reports)&155,000 breast radiology reports	
& \xmark 
&BERT	&Imporved performance\\

\cite{universeg} &Segmentation&	Images	& 53 datasets, 16 image modalities
&	\href{https://arxiv.org/pdf/2304.06131} {\cmark} 
&CNN	& Accurate segmentation \\

\cite{kim2023empirical} &Prostate MRI segmentation	&Images	&Prostate-Gland, PROSTATEx	
&  \href{https://prostatex.grand-challenge.org/} {\cmark}* 
&UniverSeg&	Can be generalized well. \\

\cite{deng2023segment} &	Segmentation	&Images	& Tissue, TCGA, and nuclei \cite{kumar2019multi}
&	 \href{https://www.cancer.gov/ccg/research/genome-sequencing/tcga} {\cmark} 
 \href{https://nucleisegmentationbenchmark.weebly.com/} {\cmark} 
&SAM \cite{SAM} &	SAM face challenges to segment dense instance objects.\\

\cite{liu2023clip}	& Segmentation (25 organs, 6 tumors)	&Images &	14 CT scans datasets 
&	 \href{https://github.com/ljwztc/CLIP-Driven-Universal-Model/tree/main?tab=readme-ov-file#dataset} {\cmark} 
&CLIP&	Promising segmentation\\

\cite{zhang2023continual}	&Segmentation (organ/tumor segmentation)&	Images	& BTCV, LiTS \cite{bilic2023liver} and JHH \cite{xia2022felix} 
&	\href{https://www.synapse.org/Synapse:syn3193805/files/} {\cmark} 
\href{https://competitions.codalab.org/competitions/17094} {\cmark} 
&CLIP	&Difficult to segment dense instance object\\

\cite{wolleb2022diffusion} &	Generating segmentation masks&	Images	&Brats2020	
& \href{https://www.med.upenn.edu/cbica/brats2020/data.html}{\cmark}
&DDPM	&Improved segmentation masks\\

\cite{rahman2023ambiguous}&Generating segmentation masks&	Images	& LIDC-IDRI, MSMRI, Bone surface segmentation	
& \href{https://wiki.cancerimagingarchive.net/}{\cmark} 
\href{https://smart-stats-tools.org/lesion-challenge-2015}{\cmark}
&AMN, ACN	&Improved segmentation masks\\

\cite{kim2022diffusion} & 	Generating segmentation masks	&Images	&XCAD 
&	\href{https://github.com/AISIGSJTU/SSVS?tab=readme-ov-file#xcad-dataset}{\cmark}
&DDPM, SPADE \cite{park2019semantic}& Improved segmentation masks \\

\cite{medsegdiff} &Segmentation	&Images	& REFUGE-2 dataset \cite{fang2022refuge2}, BraTs-2021 dataset \cite{baid2021rsna} and DDTI dataset \cite{pedraza2015open},	
&\href{https://refuge.grand-challenge.org/Home2020/}{\cmark}
\href{https://www.cancerimagingarchive.net/analysis-result/rsna-asnr-miccai-brats-2021/}{\cmark}
\href{http://cimalab.unal.edu.co/?lang=en&mod=program&id=5}{\cmark}
&DPM,  ResUNet  & Improved segmentation masks \\

\cite{medsegdiff2} & 	Segmentation&	Images &	AMOS2022, BTCV, REFUGE-2, BraTs-2021 dataset, ISIC dataset, and TNMIX dataset 
&	\href{https://amos22.grand-challenge.org/}{\cmark}
\href{https://www.synapse.org/Synapse:syn3193805/wiki/217789}{\cmark} 
\href{https://refuge.grand-challenge.org/Home2020/}{\cmark}
\href{https://www.cancerimagingarchive.net/analysis-result/rsna-asnr-miccai-brats-2021/}{\cmark}
\href{https://www.isic-archive.com/#images}{\cmark}
\href{http://cimalab.unal.edu.co/?lang=en&mod=program&id=5}{\cmark}
&DPM, UNET & Segmentation  \\

\cite{wang2023towards} & Segmentation	&Images	& LASeg dataset, Synapse dataset,  MMWHS, M\&Ms	
& \href{https://github.com/xmed-lab/GenericSSL?tab=readme-ov-file#2-data-preparation}{\cmark}
&Diff-VNet \cite{DiffUNet} &	Improved segmentation \\

\cite{fu2023recycling} & Segmentation	& Images & Abdominal CT\cite{ji2022amos}, Muscle ultrasound \cite{marzola2021deep}, Prostate MR \cite{li2023prototypical}, Brats2021 
& \href{https://github.com/mathpluscode/ImgX-DiffSeg/blob/main/imgx/datasets/README.md}{\cmark}	
&DDPM, DDIM & Improved segmentation \\

\cite{DiffUNet} & 	Segmentation& 	Images & 	BraTS2020, MSD Liver, BTCV 
&	\href{https://www.med.upenn.edu/cbica/brats2020/data.html}{\cmark},
\href{http://medicaldecathlon.com/}{\cmark} 
\href{https://www.synapse.org/Synapse:syn3193805/wiki/217789}{\cmark} 
&DDM, Diff-UNet	& Improved segmentation \\

\cite{dtan} &	Segmentation&	Images&	Kvasir-SEG \cite{jha2020kvasir}, Kvasir-Sessile \cite{jha2021comprehensive}, and the GlaS \cite{sirinukunwattana2017gland}
&	
&Diffusion model \cite{ho2020denoising}, DTAN	& Improved segmentation\\

\cite{boecking2022making}	& Classification, Segmentation &	Image-text&	MS-CXR, MIMIC-CXR	
& \href{https://physionet.org/content/mimic-cxr/2.1.0/} {\cmark}
\href{https://physionet.org/content/ms-cxr/0.1/} {\cmark}
&BERT, BioViL	& Improved performance\\

\cite{clinicalbert}	&	Report generation&	Image-text	&MIMIC-III 
&	\href{https://mimic.mit.edu/} {\cmark}
&Clinical-BERT	& Diagnosis and report generation \\

\cite{radiobert}&	Report generation	& Image-text &	OpenI 
&	\href{https://openi.nlm.nih.gov/} {\cmark}
&BERT, DistilBERT, RadioBERT &	Report generation from CXRs \\

\cite{kong2022transq} &  report generation & Image-text & IU X-ray, and MIMIC-CXR 
& 
\href{https://openi.nlm.nih.gov/} {\cmark}
\href{https://physionet.org/content/mimic-cxr/2.0.0/} {\cmark}
&TranSQ &	Medical report generation\\

\cite{tiu2022expert} &	X-ray classification & Image-text &	MIMIC-CXR, CheXpert, PadChest 
&	\href{https://www.nature.com/articles/s41551-022-00936-9#data-availability} {\cmark}

&CLIP	& SLL achieved better classification. \\

\cite{wang2022medclip} & X-ray classification &	Image-text & MIMIC-CXR, CheXpert,COVID, RSNA 
& \href{https://stanfordmlgroup.github.io/competitions/chexpert/} {\cmark}
\href{https://physionet.org/content/mimic-cxr/2.1.0/} {\cmark}
\href{https://www.rsna.org/rsnai/ai-image-challenge} {\cmark}
&BioClinicalBERT, Swin Transformer & SLL achieved better classification. \\

\cite{jang2022significantly}	& X-ray classification &	Image-text	& MIMIC-CXR, CheXpert, PadChest, Open-I, VinDr-CXR
&
\href{https://github.com/MIT-LCP/mimic-cxr} {\cmark}
\href{https://stanfordmlgroup.github.io/competitions/chexpert/} {\cmark}
\href{https://bimcv.cipf.es/bimcv-projects/padchest/} {\cmark}
\href{https://openi.nlm.nih.gov/} {\cmark}
\href{https://physionet.org/content/vindr-cxr/1.0.0/} {\cmark}
&	BERTbase, CLIP, SLIP, UniCLIP &	Improved pathology classification \\

\cite{zhang2023text}	 &	Classification &	Images &	PatchGastric
& \href{https://github.com/Yunkun-Zhang/CITE?tab=readme-ov-file#dataset} {\cmark}
&CLIP, BioLinkBERT&	Better tubular adenocarcinoma classification \\

\cite{cxrClip} &  X-ray classification	& Image-text &	MIMIC-CXR, CheXpert, ChestX-ray14, RSNA pneumonia, SIIM Pneumothorax, VinDR-CXR, Open-I 
&	\href{https://github.com/kakaobrain/cxr-clip?tab=readme-ov-file#prepare-dataset} {\cmark}
&BioClinicalBERT, CLIP, ResNet50, Swin Tiny 	& Improved pathology classification\\

\cite{gloria}&	Image-text retrieval , Classification, Segmentation &	Image-text	& CheXpert, RSNA Pneumonia, SIIM Pneumothorax	
& \href{https://github.com/marshuang80/gloria/blob/main/gloria/datasets/image_dataset.py} {\cmark} 
&BioClinicalBERT, ResNet &	Improved performance\\

\cite{yang2023diffmic} & Classification	& Images &	PMG2000, HAM10000, APTOS2019 
& 	\href{https://github.com/scott-yjyang/DiffMIC?tab=readme-ov-file#datasets}{\cmark}
&DDPM, DCG model &	Improved performance \\


\cite{adaptivediffusion}	& MRI reconstruction &	Images &	fastMRI, IXI 
&	
\href{http://brain-development.org/ixi-dataset}{\cmark}
\href{https://fastmri.med.nyu.edu/}{\cmark}
&AdaDiff &	Improved reconstruction performance.\\

\cite{ozbey2023unsupervised} & 	MRI-CT  translation &	Images &	IXI, BRATS, MRICT
&	
&SynDiff	& Improved performance\\

\cite{zhu2023make} & Modality translation &	Images	& S2M, RIRE , 
& \href{https://rire.insight-journal.org/index.html}{\cmark}
&Make-A-Volume &	 Improved performance \\

\cite{mao2023disc}&  MRI Super-Resolution &	Images &	IXI, in-house clinical brain MRI dataset 
& \href{https://github.com/Yebulabula/DisC-Diff?tab=readme-ov-file#dataset--pretrained-models}{\cmark}
& DisC-Diff   &Improved performance \\

\cite{wang2023inversesr} & 	MRI Super-Resolution &	Images &	IXI, UK Biobank.
& 
\href{https://brain-development.org/ixi-dataset/}{\cmark}
\href{https://www.ukbiobank.ac.uk/}{\cmark}
&	InverseSR&	Improved performance\\
\end{longtable}
 \raggedright * Prostate-Gland is not available \\ 
\endgroup
\twocolumn

\section{Discussions}
\label{discussions}

This section highlights key findings and insights, including brief taxonomies and results, opportunities provided by FMs, and challenges to FMs in healthcare. Additionally, the lessons learned provide valuable guidance for future work, which is discussed along with potential directions for advancing the field.



\subsection{Taxonomies and Comparative Results}
\label{results}

This section briefly outlines the taxonomy of FMs utilized for healthcare applications. FMs are powerful tools in various domains including healthcare that revolutionized medical practice and research. 
The progression of medical FMs for various healthcare applications is represented in Figure~\ref{fig:fig3}. We categorized the model into five different categories: CLLMs, Omics, Vision-Based Medical FMs, Video/Audio FMs, and Multimodal medical FMs. CLLMs, such as BioBERT and Clinical-BERT, were on the first BERT-based models. More recent CLLMs includes GatorTronGPT \cite{GatorTrongpt}, BioGPT \cite{biogpt}, MedPaLM2 \cite{medpalm2}, Me LLaMA \cite{xie2024me}, and BioMistral \cite{labrak2024biomistral}. Medical FMs that utilized omics data (Section~\ref{omics}) emerged later, leveraging transformers to analyze DNA, RNA, and protein sequences, aiding in genomics and molecular biology research. Vision-Based Medical FMs like Med3D and VisualGPT adapted image-based architectures for radiology and medical imaging, assisting in interpreting X-rays and MRIs. Video/Audio Models, exemplified by Whisper and AudioGPT, enable medical transcription and audio analysis, contributing to telemedicine and patient monitoring. Multimodal FMs represent the most recent and advanced developments, integrating text, image, and audio processing capabilities for more comprehensive healthcare applications. These models combine diverse model architectures to enable cross-modal understanding, such as interpreting imaging data alongside clinical notes. Some of the recent multimodal MFMs includes, Macaw-LLM \cite{lyu2023macaw}, OneLLM \cite{han2024onellm}, and SurgVLP \cite{yuan2023learning}.
Most of the FMs are based on transformer architecture, which has been widely adopted after the great success since 2017 \cite{vaswani2017attention}. BERT was one of the first FMs designed for sequence modeling. FMs got attention after ChatGPT among the general public, leading to the development of similar models such as Bard, Chinese GPT, etc. 
FMs have revolutionized healthcare to assist clinical decision support, medical image analysis, drug discovery, and personalized medicine. Their ability to process large volumes of heterogeneous data, extract meaningful patterns, and generate insights has empowered healthcare professionals to improve diagnostic precision, personalized treatment strategies, and patient outcomes. For instance, FMs have been instrumental in developing medical chatbots, generating clinical reports, extracting information from medical texts, and facilitating language translation to enhance communication and information exchange in healthcare settings (Section~\ref{NLP}).
FMs also contributed to computer vision to facilitate various tasks including image generation, classification, segmentation, and modality translation (Section~\ref{computervision}). FMs aid in reconstructing high-resolution medical images, enhancing image quality, and synthesizing missing or corrupted modalities. Moreover, such models have been integrated into augmented reality systems to enable real-time guidance and contextual information during medical interventions, training, and diagnostic procedures.

FMs in healthcare perform well on various tasks, including NER, biomedical QA, image segmentation, and protein structure prediction.  While general-purpose models (BERT, GPT-3, ChatGPT, SAM) can be applied to medical tasks, specialized models trained on medical-specific datasets (e.g., GatorTronGPT, BioMegatron, MedSAM) tend to outperform their general counterparts in multiple tasks including clinical NLP and medical image segmentation. Furthermore, compared to traditional CLIP, MedCLIP showed strong zero-shot capabilities in pairing medical images with clinical reports, enabling efficient disease diagnosis and image-text interaction. 
FMs in healthcare are diverse, encompassing various architectures and models designed for different medical tasks and data types. The key architectural types include encoder models like BERT and BioBERT, which excel in NLP tasks such as classification and sequence labeling. Decoder models like GPT and GatorTronGPT for text generation and report creation, and encoder-decoder models like T5, BART, and bioMegaTron are useful for tasks like summarization and translation. These models power applications in clinical NLP, medical computer vision, and omics, with each architecture tailored for specific tasks such as NER, medical image segmentation, and gene expression prediction. 
Larger models tend to perform better in complex tasks but are also more resource-intensive, requiring significant computational power and memory. For instance, GatorTronGPT (20 billion parameters) demonstrated superior performance than BioBERT (345 million parameters) in clinical NLP tasks (e.g., NER, QA, RE) by leveraging much larger datasets, including 82 billion words from de-identified clinical texts. Its ability to handle biomedical question-answering and text similarity tasks is unmatched by smaller models. Vision-based models (e.g., MedSAM) also excel in medical image segmentation, outperforming traditional models (e.g. U-Net). However, generalization across domains remains a challenge; while models trained on large general-purpose datasets can perform a wide range of tasks, domain-specific models like BioBERT often require fine-tuning to achieve optimal results in specialized areas like omics or image analysis. 


\begin{figure}[htbp]
    \centering
    \includegraphics[width=\columnwidth]{figures/Fig3.pdf}
    \caption{Taxonomy of medical FMs evolved in a variety of complex healthcare data and tasks. Each medical FMs is represented in different color by on diverse base model architectures — such as BERT, Transformers, GPT, CNN, and LLaMA — each tailored to the unique demands of different medical tasks.}
    \label{fig:fig3}
\end{figure}

\subsection{Opportunities}
\label{Opportunities}
FMs present significant opportunities to advance healthcare by addressing key challenges and enhancing various aspects of medical practice. In this section, we explore crucial opportunities provided by medical FMs.

\textbf{Small Data Size in Rare Diseases:}
Medical FMs can learn robust representations from general medical data, which can then be fine-tuned on small datasets where obtaining datasets is not feasible, especially related to specific rare diseases. FMs can adapt to new domains with minimal data using few shots or zero short learning (e.g. RetiZero\cite{wang2024common}), enabling accurate disease characterization, prognosis, and treatment recommendations even for rare conditions. Furthermore, multimodal FMs can benefit from cross-domain learning, where insights from one modality can enhance understanding in another. For instance, a model trained on common disease imaging data can be fine-tuned on limited imaging data for a rare condition, improving diagnosis accuracy.

\textbf{Robustness in Medical Foundation Models:} Medical FMs are pre-trained on diverse datasets encompassing a wide range of patient data, exposing the model to variations that it might encounter in real-world settings and helping it learn more generalized representations. Furthermore, incorporating multiple data types—such as text from clinical notes, lab results, and radiology images—medical foundation models develop a richer understanding of patient health and improve robustness. Moreover, SLL techniques allow medical FMs to learn from raw, unlabeled data to form a deep contextual understanding and provide robust prediction across multiple healthcare applications. Robustness is further enhanced through fine-tuning on specific datasets, such as data from a particular hospital or region.

\textbf{Data Privacy Preservation:} Medical FMs pre-trained on vast amounts of generalized data can achieve accurate performance on new medical tasks with limited fine-tuning data. It significantly reduces the exposure or process of sensitive patient information. Furthermore, since medical FMs can rely on SSL techniques, it reduces the need for manually annotated patient records, further preserving privacy and avoiding bias. Furthermore, such models can leverage synthetic data from large public datasets, which helps to simulate rare cases or specific patient characteristics without involving real patient data \cite{GatorTron, GatorTrongpt}. It enables models to generalize better across populations without having to access large, private datasets.

\textbf{Trustworthiness:}
Empowering clinicians with tools and interfaces to intervene in the model's decision-making process will enable personalized and context-aware healthcare solutions. For instance, clinicians can use FM-generated recommendations based on patient-specific factors, clinical guidelines, and individual preferences to enhance the relevance and utility of FM outputs in clinical decision-making. Furthermore, FMs can be fine-tuned and customized to meet specific clinical requirements and references to facilitates transparency for clinicians to understand the underlying reasoning of FMs. This will foster trust and confidence in their usage in medical practice.

\textbf{Flexibility and Applicability:}
FMs to adjust to diverse medical data, disease types, and healthcare contexts can be termed as flexibility.
FMs are trained on large amounts of diverse medical datasets, including medical literature, patient records, and clinical guidelines. FMs can be tailored and fine-tuned to accommodate various data modalities, disease phenotypes, and healthcare contexts. In practice, FMs can also provide real-time assistance through diagnosis, treatment planning, and decision-making. Harnessing such flexible models will reduce the need for task-specific models and aid healthcare organizations in developing customized and context-aware solutions that align with specific clinical needs and operational requirements.
Furthermore, applicability refers to the ability of FMs in addressing a wide range of healthcare taks and use cases. FMs can be applied to diverse tasks with limited fine-tuning such as clinical decision support, medical image analysis, predictive modeling, natural language understanding, drug discovery, and precision medicine. Moreover, the applicability of FMs extends beyond traditional healthcare domains to encompass emerging areas such as digital health, telemedicine, population health management, and patient engagement. FMs are opening up new frontiers for innovation and transformation in healthcare delivery and patient care.

\subsection{Challenges to FM in Healthcare}
\label{challenges}

\textbf{Cost}:
FMs are experiencing rapid growth with training parameters increasing significantly compared to traditional models. The models' increasing size demands computationally expensive hardware requirements. This poses a handful of challenges in terms of the training cost, inference cost, debugging, monitoring, and maintenance in production applications \cite{bommasani2021opportunities}.


\textbf{Interpretability:} 
Interpretable FMs in healthcare present significant challenges due to the complexity and variety of tasks these models are trained to perform. Unlike task-specific models, FMs are trained on extensive and diverse datasets so they can be adapted to various downstream tasks \cite{bommasani2021opportunities}. However, this adaptability complicates the ability to fully understand or predict the model’s behavior in critical medical contexts. For instance, models like GPT-3, trained for general language tasks, can exhibit unexpected behaviors in domain-specific healthcare tasks and unsuitable for medical advice such as medical dosages, emergency, and abortion, without providing transparent insight into the underlying mechanisms \cite{lechner2023challenges}.
Furthermore, due to the heterogeneous nature of healthcare data encompassing patient histories, imaging, lab results, and genetic information, it is challenging for FMs trained on such diverse inputs to produce predictions that are easy to interpret, as the decision-making process may involve a combination of factors that are not easily explainable.
Moreover, in healthcare, understanding the causal relationships behind predictions is critical. General interpretability techniques, such as attention mechanisms or feature importance rankings, often only provide surface-level insights (i.e., correlations rather than causations) \cite{lyu2024language}.  Causal methods, such as counterfactual interventions, allow clinicians to see how changes in patient data would affect diagnoses or treatments, ensuring decisions are based on biological mechanisms rather than mere correlations \cite{feder2021causalm, creswell2022selection}. However, building reliable causal models is challenging due to their computational complexity and difficulty in generalizing across diverse medical tasks.

\textbf{Validation:}
Validating the performance and efficiency of FMs in healthcare presents a significant challenge due to the complex and dynamic nature of medical data. Traditional validation approaches, such as cross-validation and holdout validation, may not be sufficient to assess the reliability and generalizability of FM outputs in real-world clinical settings. Moreover, the lack of standardized evaluation metrics and benchmarks tailored to healthcare applications complicates the validation process. One of the key challenges in validation is ensuring the robustness and stability of FM across diverse patient populations, disease cohorts, and clinical scenarios. Data quality, disease prevalence, and patient demographics can impact the generalization ability of FMs. This leads to biases, over-fitting, or underperformance in specific contexts.

\textbf{Verification:}
Verification of FMs in healthcare involves ensuring the correctness, reliability, and safety of model behavior and predictions. It involves various aspects such as model design, implementation, training, testing, and monitoring, to mitigate risks associated with erroneous or biased decision-making.
One of the primary challenges in verification is validating the accuracy and consistency of FM outputs across different stages of the model life cycle. Ensuring reproducibility and consistency in model training, hyperparameter tuning, and optimization processes is essential to verify the reliability of FM-generated predictions. Additionally, verifying the robustness of FMs against adversarial attacks, data perturbations, and distribution shifts is critical to assess their resilience in real-world healthcare scenarios. Furthermore, verification should address concerns related to data privacy, security, informed consent, and algorithmic fairness to mitigate potential risks associated with FM usage.

\textbf{Privacy and Security:}
Security and privacy considerations include protecting user data confidentiality, ensuring model integrity, and maintaining system availability. The possible threats to FMs include inference attacks, adversarial examples, and data poisoning \cite{bommasani2021opportunities}. Furthermore, since the FMs are adopted for various downstream tasks, they are prone to be compromised as a single point of failure. For instance, injecting malicious data as a training sample could impact all the dependent downstream applications.

\textbf{Social bias:}
Social bias involves an uneven preference or bias towards one entity, individual, or group over another \cite{lee2023survey}. Unfairness arises when individuals are not treated equally in any tasks based on factors such as gender, race, age, and marital status \cite{lee2023survey}. FMs may result in social bias such as significant social biases in Stable Diffusion \cite{saravanan2023exploring}, particularly in gender and racial stereotypes.

\textbf{Scale:}
The volume of training data for FM development is expanding annually. This trend is expected to persist due to the reliance of models on publicly available datasets \cite{kaplan2020scaling}. However, publicly available data is relatively smaller compared to industrial models \cite{bommasani2021opportunities}. Therefore, there is a need for models trained in large amount of data that are capable of handling the complexities associated with multi-modal FM datasets \cite{zhuang2023foundation}.

\textbf{Legal and ethical consideration}
Legal and ethical considerations play a vital role in navigating the challenges posed by the deployment of FMs. From a legal perspective, issues arise on data collection methods, copyright laws, and privacy regulations. Furthermore, regulations such as the General Data Protection Regulation (GDPR) influence the acquisition and utilization of training datasets. Another issue arises from the legal responsibility of model creators in cases when a model generates harmful responses, especially in healthcare \cite{seyyed2021underdiagnosis}. Moreover, legal protections for model outputs may result in issues associated with free speech, ownership rights, and copyright laws. 
In addition to the legal framework, ethical considerations are also important. Major ethical challenges associated with the development of FMs include obtaining informed consent for data usage, ensuring safety and transparency, addressing algorithmic fairness and biases, and safeguarding data privacy.

\textbf{Accuracy/Performance}
In healthcare, achieving high performance is important for the successful implementation of AI models. FMs present a promising foundation by being pre-trained on extensive and diverse datasets. However, they may not meet the specific accuracy requirements of healthcare especially due to hallucination \footnote{Hallucination refers to the generation of plausible but incorrect or unverified information.}. Fine-tuning these models with domain-specific training data is a common strategy to improve accuracy. However, this process can be resource-intensive. Striking a balance between accuracy improvement and resource allocation is crucial for effectively leveraging FMs in healthcare settings.

\subsection{Open Research Issues and Lessons Learned}
\label{openresearch}

\paragraph{Tailoring FMs for Healthcare}

General FMs like BERT and GPT-3 struggle to perform well when applied directly to healthcare tasks because medical data—such as clinical notes, radiology reports, and biomedical research— is complex and different from web text data. Therefore, models such as BioBERT and GatorTron are specialized variants of BERT, fine-tuned on large-scale biomedical datasets such as PubMed abstracts. These models outperform general models in medical NLP tasks like NER and QA because they are adapted to the nuances of biomedical text \cite{GatorTron}. Hence, FMs should be carefully fine-tuned and adapted for the medical domain to be effective and not directly ported from general domains without specialization \cite{soroush2024large}. 
However, even with finetuning, these FMs remain constrained by the limitation of their underlying architectures. In both NLP and imaging tasks, current open-source FMs lag behind SOTA closed-source models, which depreciates the progress in medical research and clinical application, when they are simply adapted from pretrained FMs. Unlike general domains, clinical decision making, patient interactions, and diagnosis not only demand specialized knowledge, but also have unique data patterns. Consequently, training methods and model architectures that are specifically designed towards clinical tasks are required to overcome the inherent limitation of current backbone models in healthcare application.  

\paragraph{Self-Supervised Learning and High-Quality Medical Data}

SSL is a key enabler of progress in healthcare FMs. It helps overcome the bottleneck of limited labeled data, enabling the creation of scalable and task-agnostic models that can be fine-tuned for specific medical tasks. For instance, GatorTronGPT utilizes SSL and leverages large-scale, unlabeled clinical data from EHRs that can be used for various downstream tasks within healthcare applications. While SSL helps mitigate the issue of the availability of high-quality, annotated medical data, many healthcare-specific tasks still require annotated datasets for fine-tuning. However, developing FMs for healthcare requires the availability of high-quality, annotated medical data \cite{zhang2024data,sun2024medical}. Due to privacy concerns and the complexity of medical data, obtaining and sharing this data remains difficult. Creating high-quality, labeled medical datasets is essential for advancing the capabilities of FMs in healthcare . Furthermore, the development of privacy-preserving techniques such as federated learning could help address data-sharing challenges by enabling model training on distributed data without compromising patient confidentiality \cite{he2024foundation,sun2024medical}.

\paragraph{Multi-modality and Artificial General Intelligence}

Traditional models that focus on a single modality are limited in their ability to provide comprehensive healthcare insights. However, medical tasks often require the integration of diverse data types, including text, images, and even omics data. Recent advancements in SOTA models, such as GPT-4, underscore the significance of developing models capable of processing multi-modal inputs. GPT-4 can interpret visual inputs and deduce and understand the characteristics based on user-provided text prompts. This trend of integrating multi-modal data has also been observed in recent medical AI research \cite{sun2024medical}. The success of MedSAM and MedCLIP demonstrates the effectiveness of multi-modal approaches in various tasks such as diagnostic image classification and report generation. Therefore, for effective healthcare applications, multi-modality models that integrate data from various sources (e.g., images, clinical text, genomics) provide a more holistic view and better support for complex medical decision-making. Furthermore, Artificial General Intelligence (AGI) in healthcare has the potential to revolutionize patient care by integrating advanced models that can understand and analyze complex clinical data across multiple modalities, such as text, images, and audio \cite{li2023artificial}. However, implementing AGI in healthcare presents challenges, including ethical concerns, data privacy, and the need for large-scale, high-quality datasets, requiring careful oversight and collaboration between AI experts and medical professionals\cite{li2024artificial}.

\paragraph{Model Interpretability for Clinical Adoption}

One of the significant challenges that emerged is the "black-box" nature of many FMs, which limits their adoption in healthcare, where clinicians require transparency to trust automated systems \cite{fu2024championing}.  Despite achieving notable performance in NER, medical Q\&A, and biomedical relation extraction, especially when these models were finetuned, these models often lack interpretability in their decision-making process.  For instance, the CLIP model's sensitivity to red circles highlights how such biases may result in misclassifications \cite{shtedritski2023does}. This lack of transparency undermines the trustworthiness of medical AI and depreciates its integration in the clinical setting. Additionally, the current evaluation of medical AI systems is predominantly focused on computational feasibility. However, assessing interpretability solely from a computational perspective may lead to a misalignment between the interpretability researchers consider optimal and the interpretability that meets the actual needs of patients. Therefore, AI models require interpretability and explainability in learning algorithm and their model design in a human-centered way for wide clinical adoption \cite{chen2022explainable}.

\paragraph{Model Size and Practical Use}

Although larger models demonstrate superior performance, they are also more computationally expensive and difficult to deploy in clinical environments, especially where resources are limited. For instance, the MedSAM model was trained on 20 Nvidia A100 GPUs with total GPU memory of 1600 GB, and BiomedCLIP was trained on 16 Nvidia A100 GPUs with total GPU memory of 640 GB.Such computational demands are beyond the reach of many researchers, potentially stalling innovation in the field. Therefore, the balance between performance and computational cost needs careful consideration, particularly for deployment in real-time clinical settings where resource constraints are a concern. Scalable and efficient models could be easily applied in a range of healthcare settings, from large hospitals to smaller clinics \cite{chen2023harnessing}.
 
\paragraph{Personalized Medicine}
The use of FMs is showing great promise in personalized medicine \cite{chen2023survey}. By analyzing large datasets that include genomic data, FMs can uncover previously unknown relationships between genes and diseases, helping to predict disease risk and treatment response. 
Models like scGPT and GeneGPT are trained on vast amounts of genomic and single-cell data to provide insights into patient-specific treatments. These models allow for personalized recommendations in areas like drug discovery and cancer treatment, where traditional one-size-fits-all approaches are ineffective. FMs will play a central role in advancing precision medicine. While these FMs, leveraging vast knowledge acquired from the biomedical research papers and articles, have reached expert-level proficiency in certain domains — offering novel insights for evidence-based treatment plans and personalized diagnoses — they continue to struggle with making clinician-level decisions, such as determining the most effective treatment for individual patients. Therefore, they must be carefully validated to ensure that their predictions are reliable and actionable 

 \paragraph{Privacy-preserving Techniques} Since most of the healthcare data is not publicly available, solving privacy issues of patient privacy and implementing federated learning will provide FMs with a large amount of training data \cite{jia202310}. Furthermore, it will enable collaborative model training across multiple institutions while ensuring data security and confidentiality.

\paragraph{Robust FMs}FMs that are susceptible to adversarial attacks may result in wrong information
and diagnosis that leads to incorrect treatments or delays in necessary interventions. Moreover, adversarial attacks can compromise patient privacy and may result in severe breaches of confidentiality \cite{khalid2023privacy}. This would result in raising ethical and legal concerns. Therefore, there is still a need to incorporate robust FMs before deploying them into clinical settings.

\paragraph{Multi-purpose FMs and Cost-effective Deployment}FMs are generally multi-purpose; however, due to the complex structure of healthcare data, general-purpose medical FMs are not yet explored that handle various applications. Designing a universal large-scale FM tailored specifically for healthcare would facilitate the creation of a centralized model that could be adopted across various healthcare domains. Furthermore, developing efficient training and inference algorithms and exploring cloud-based solutions will facilitate cost-effective deployment in healthcare settings \cite{gao2024cost, xu2024survey}.




\section{Conclusion}
\label{conclusion}

We conducted a detailed survey to review the recent advances of FMs in the healthcare domain. Our study explained the wide range of potential applications and significant contributions of FMs in revolutionizing various aspects of healthcare delivery. We provided insights into both the key general-purpose FMs and specialized models designed for healthcare applications. Moreover, our survey explored nearly all healthcare applications associated with FMs, ranging from clinical decision support systems to medical image analysis and drug discovery. By explaining the detailed landscape of the medical domain, our survey fills a crucial gap in existing literature and serves as a valuable resource for researchers and practitioners in the field.

\appendices

\section*{Acknowledgment}
We sincerely appreciate Chintan Acharya for his invaluable assistance in creating the figures for this work. We also extend our sincere gratitude to the anonymous reviewers for their insightful comments and suggestions.


\section{Background}
\label{background}
\bstctlcite{IEEEexample:BSTcontrol}

\subsection{Transformer}

A Transformer \cite{vaswani2017attention} is a neural network architecture initially developed for sequence modeling. It addressed the inefficient computation and poor performance of RNNs caused by the sequential processing of the information. Due to the attention mechanism in transformers, sequences can be processed in parallel, which is faster and more efficient during training.

In a transformer, the attention mechanism is computed by calculating a weighted sum of all token representations in a sequence. These weights are determined based on the similarity between the tokens. Since there is no convolution or recurrence, positional encoding is used to represent the relative positions of tokens in a sequence.

\begin{figure}[t]
    \centering
    \includegraphics[width=0.95\columnwidth, trim = 0cm 2cm 0cm 1cm, clip]{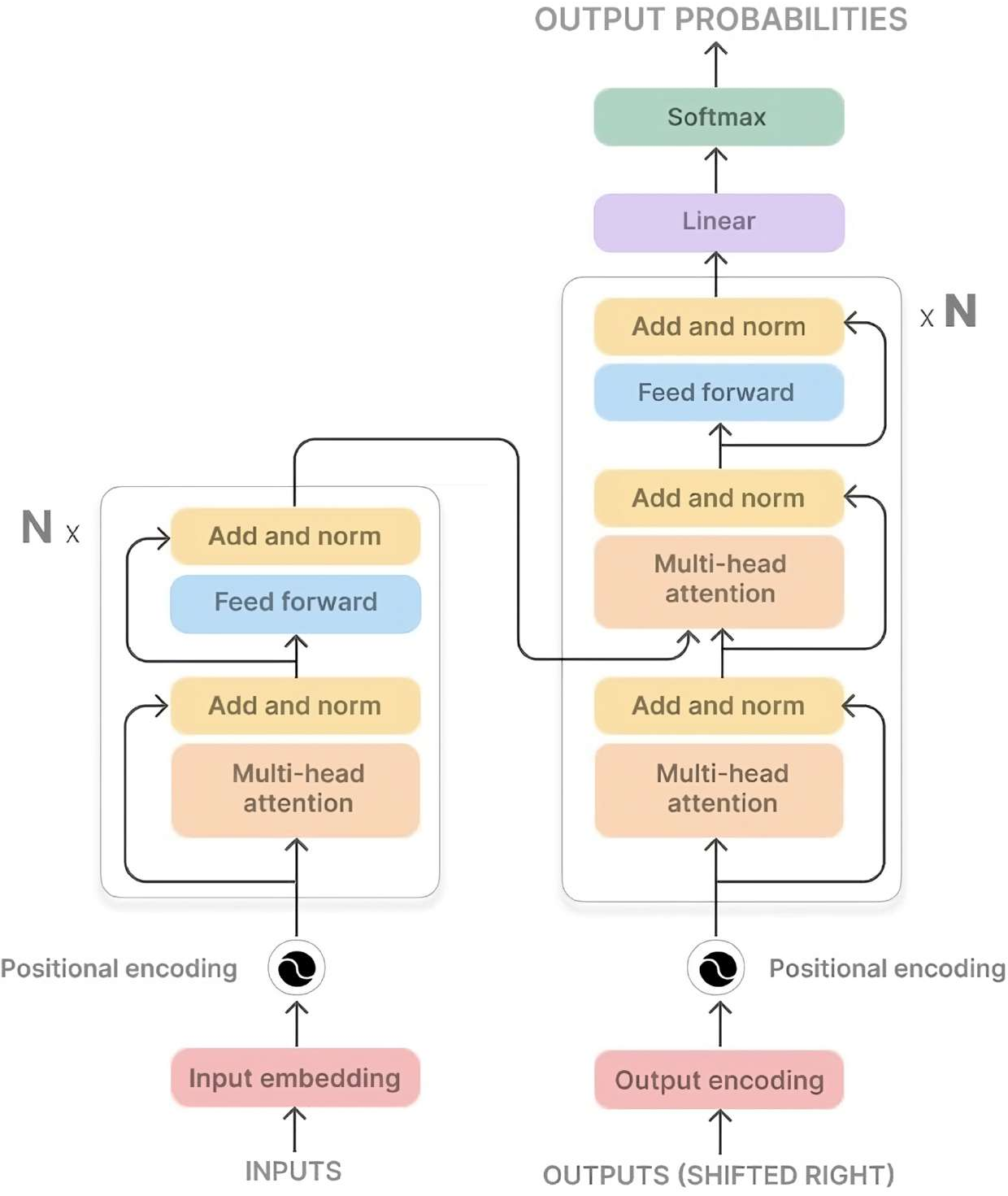}
    \caption{Transformer architecture \cite{vaswani2017attention} employs an encoder-decoder structure with multiple stacked layers (Nx), each containing multi-head self-attention and feed-forward layers.}
    \label{fig:transformer}
\end{figure}

The Transformer architecture shown in Figure~\ref{fig:transformer} uses an encoder-decoder framework, where the encoder processes the input into intermediate representations using attention blocks, and the decoder utilizes these representations to generate the output, with attention mechanisms enabling the model to focus on relevant parts of the input at each stage.
The positional encoding vectors are added to the embedding vectors for the tokens at the input to the self-attention mechanism. This helps with learning long-range dependencies in a sequence by attending to the positional encoding vectors of the tokens. Transformers have led to the development of SOTA models including BERT \cite{devlin2018bert}, GPT \cite{gpt}, and other models \cite{awais2023foundational}.
Transformers were initially designed for NLP tasks but have subsequently been expanded to various other domains, including computer vision. Vision transformer (ViT), first proposed in 2021 \cite{ViT} are a type of transformer specifically designed for computer vision tasks. It converts an image into a sequence of patches, and then it utilizes a self-attention mechanism to calculate meaningful embeddings. The positions of patch embeddings are retained using positional encoding.

\subsection{Attention}
\label{Attention}

\begin{figure}[t]
    \centering
    \includegraphics[width=0.95\columnwidth, trim = 0cm 7cm 0cm 3cm, clip]{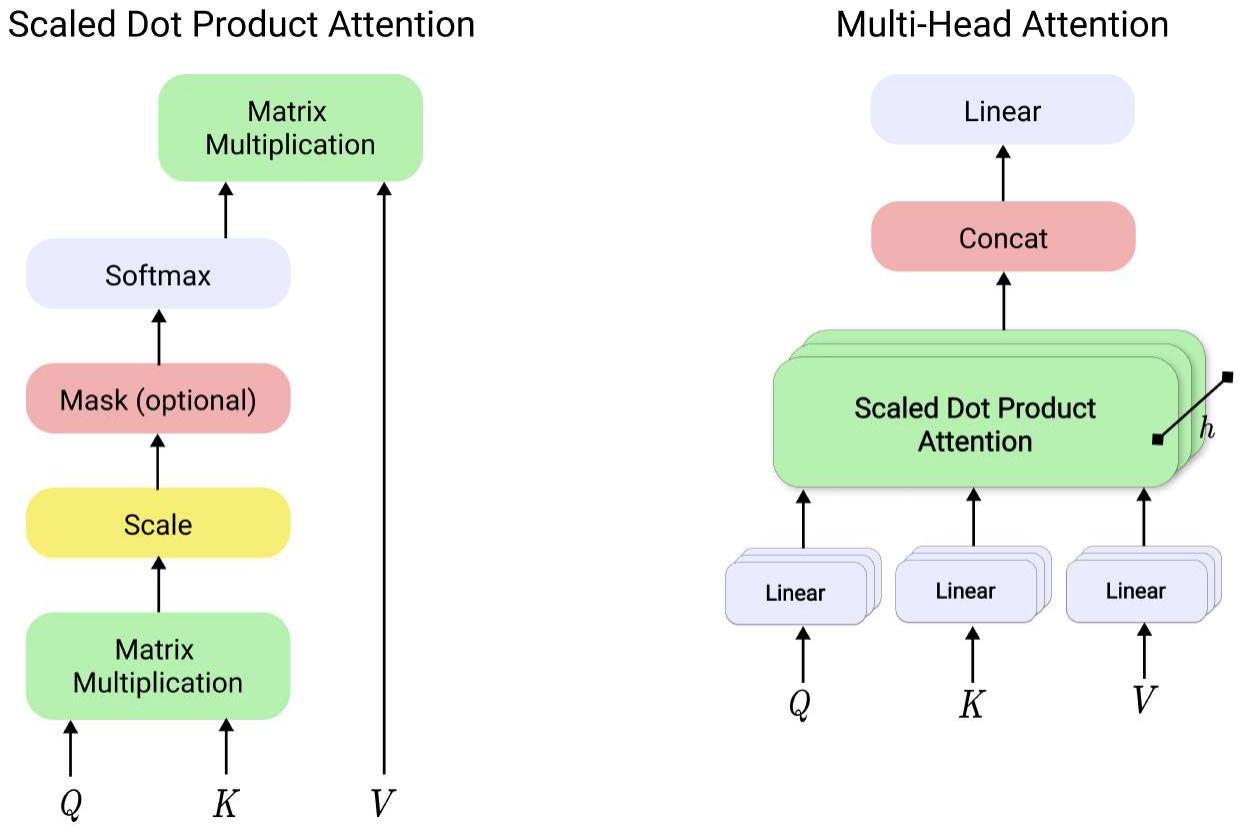}
    \caption{Attention mechanism in Transformer architecture.}
    \label{fig:self}
\end{figure}

Attention mechanism in deep learning originated from the human visual attention that selectively focuses on one or more aspects of information within a vast array of available data \cite{niu2021review}. For instance, Itti et al. \cite{itti1998model} proposed visual attention, inspired by the primate visual system, to identify salient regions. The attention mechanism has been used mainly with RNNs or CNNs for multiple tasks \cite{niu2021review}. However, transformers rely solely on attention, unlike CNNs and RNNs which use attention in part of the model. The similarity between tokens in a sequence is computed by the attention mechanism in transformers. Multiple attention mechanisms have been proposed recently, including self-attention \cite{vaswani2017attention}, multi-head attention \cite{vaswani2017attention}, masked attention \cite{cheng2022masked}, and criss-cross attention mechanisms  \cite{huang2019ccnet}.
In self-attention represented in Figure~\ref{fig:self}, the input sequence in each element is associated with query, key, and value vectors. The model then computes attention weights by measuring the similarity between the query and key vectors for all elements in the sequence. These weights guide the model in taking a weighted sum of the value vectors for each element, effectively enabling the model to selectively focus on different parts of the input when processing each element.
Multi-head attention learns multiple representations of the input sequence, each of which can attend to different aspects of the sequence. This is done by performing multiple self-attention operations in parallel, on different input representations. Cheng et al. \cite{cheng2022masked} proposed Mask2Former which utilizes masked attention in the transformer decoder. Masked attention restricts attention to specific regions of interest, such as the foreground region of a predicted mask, during the calculation of attention scores. Criss-cross attention \cite{huang2019ccnet} efficiently captures full-image contextual information for tasks like semantic segmentation and object detection. It utilizes a novel Criss-Cross Network with a criss-cross attention module, allowing each pixel to gather contextual information from its surroundings.

\subsection{Self-supervised learning (SSL)}
\label{SSL}
\begin{figure}[t]
    \centering
    \includegraphics[width=0.95\columnwidth, trim = 0cm 6cm 0cm 2cm, clip]{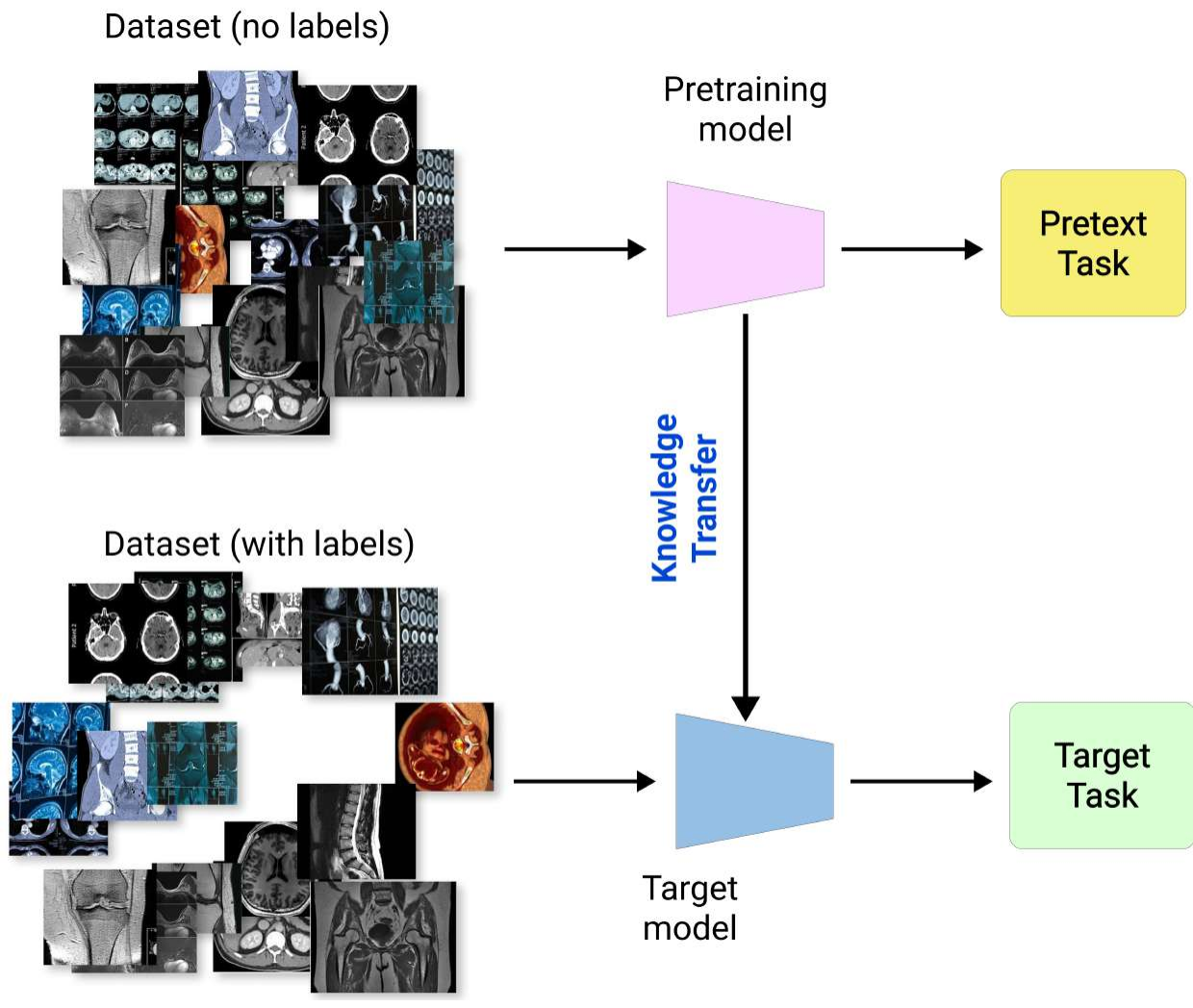}
    \caption{Representation of self-supervised learning where an unlabeled dataset is used to create a pretext task for learning representations, which are then transferred to a labeled dataset to perform a target task.}
    \label{fig:SSL}
\end{figure}

The limitation of supervised learning is that it depends heavily on large amounts of labeled data, which is a time-consuming, expensive, and often infeasible task \cite{huang2023self,liu2021self}. Furthermore, since the data are annotated based on the labels due to correlation, it introduces generalization error \cite{huang2023self} and is prone to adversarial attacks \cite{liu2021self}. In contrast, SSL can be utilized on extensive datasets without requiring labels. SLL utilizes the inherent structure or context within the data itself to create meaningful representations. By learning from the structure of the data itself, SSL reduces the reliance on labeled datasets and improves generalization to downstream tasks (Figure~\ref{fig:SSL}. SSL can be divided into three main categories: contrastive, generative, and adversarial learning \cite{liu2021self}. Contrastive learning algorithms train models to distinguish between distinct data points based on similarity within the representation space. Generative learning algorithms train models to generate realistic data. Adversarial learning is a hybrid generative-contrastive learning approach that improves representations by reconstructing the original data distribution rather than individual data samples, using discriminative loss functions as objectives.

\subsection{Human feedback reinforcement learning}
Reinforcement learning (RL) is an ML-based algorithm where an agent engages with an environment and makes a series of decisions to maximize cumulative rewards \cite{macglashan2017interactive}. It can be formulated using a Markov Decision Process (MDP), mathematically, $\langle S, A, T, R, \gamma \rangle$ \cite{macglashan2017interactive, arumugam2019deep}, where $S$ represents multiple states in an environment and selects actions ($A$). The transition probability ($T$) specifies how the agent moves from one state ($s$) to another ($s'$), and a reward function ($R$) provides feedback on the actions taken. The discount factor ($\gamma$) is a critical parameter that influences the relative importance of future discounted rewards in the decision-making process. The agent's goal is to find a policy ($\pi: S \rightarrow A$) that specifies how to select actions in each state to maximize the expected reward. Although RL learns complex behavior with predefined reward functions, however, it can be challenging to express some behaviors as rewards \cite{arumugam2019deep}. Therefore, human feedback RL (HFRL) is when the agent learns about their action from feedback from a human trainer as shown in Figure~\ref{fig:RLHF}. The human feedback has resulted in significant performance improvements \cite{nakano2021webgpt}.

\begin{figure}[t]
    \centering
    \includegraphics[width=0.95\columnwidth]{figures/background/FigS4.pdf}
    \caption{Representation of Human feedback reinforcement learning.}
    \label{fig:RLHF}
\end{figure}

\subsection{CLIP}
\label{CLIP}
Traditional DL algorithms have demonstrated promising performance in multiple computer vision tasks. However, they rely heavily on extensive labeled training data, which is labor-intensive and expensive. Additionally, these algorithms struggle to generalize beyond the specific tasks they are trained for \cite{CLIP}. These challenges are addressed by introducing CLIP \cite{CLIP}, a novel neural network-based architecture that leverages natural language-supervised vision data readily available on the internet, effectively reducing the dependence on expensive labeled datasets while enabling broader generalization.

\begin{figure}[t]
    \centering
    \includegraphics[width=0.95\columnwidth, trim = 0cm 5cm 0cm 5cm, clip]{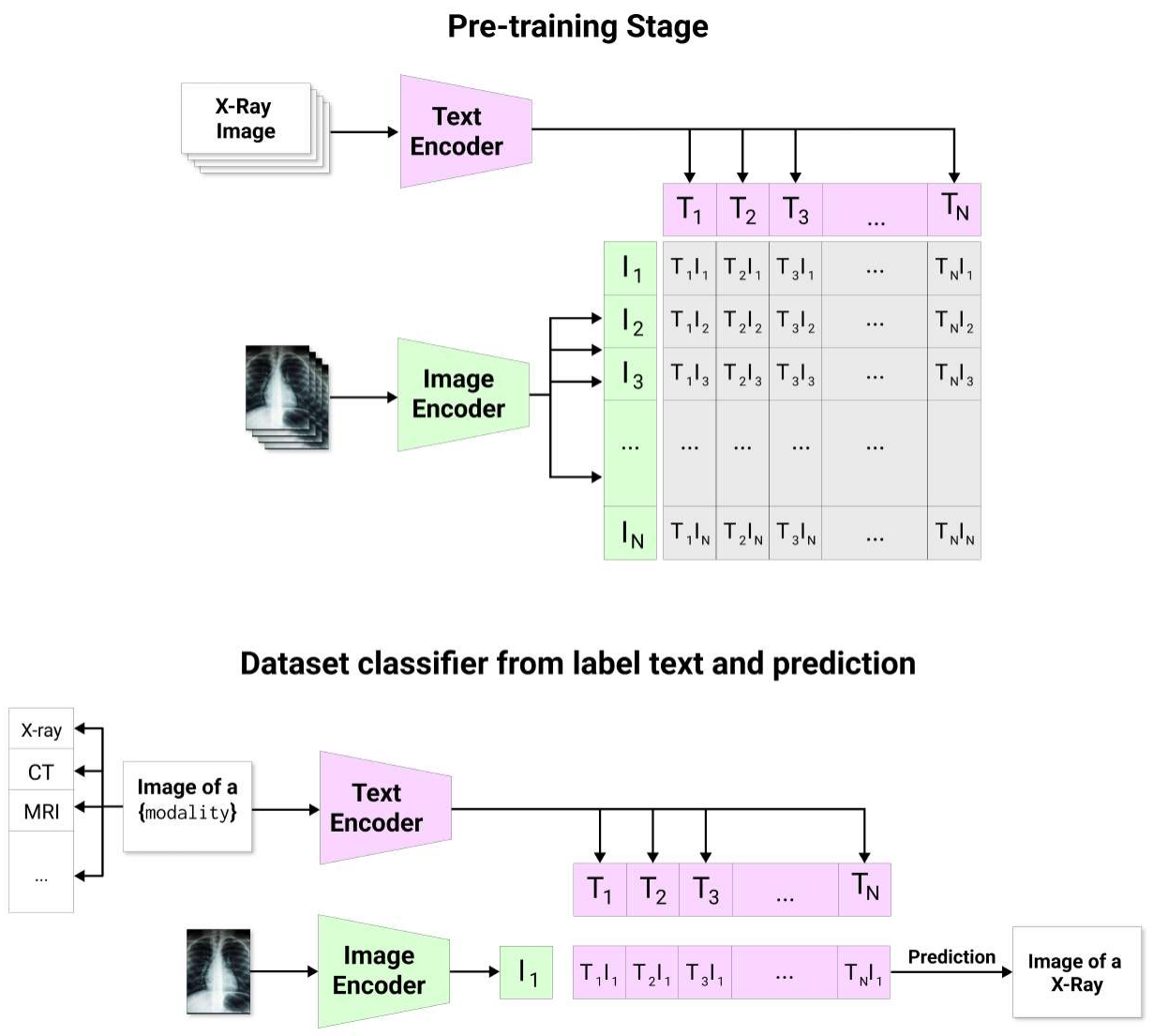}
    \caption{Contrastive Language-Image Pre-Training (CLIP) model uses a dual-encoder to align text and image embeddings through contrastive learning, enabling zero-shot transfer to new visual tasks.}
    \label{fig:CLIP}
\end{figure}

CLIP is a robust neural network architecture with zero-shot capabilities \cite{gpt2} that achieves various classification benchmarks using natural language instructions, without direct optimization for each specific benchmark \cite{gpt2}. It is a simplified variant of ConVIRT \cite{zhang2022contrastive} and is trained on a dataset called WebImageText, comprising 400 million pairs of images and corresponding text collected from the internet. Multiple works such as \cite{zhang2022contrastive, gomez2017self, joulin2016learning, desai2021virtex} utilized text-image representations, however, these methods operate under supervised, weakly-supervised, self-supervised, or unsupervised learning paradigms. In contrast, 
CLIP utilizes natural language as its training signal. The key advantage of leveraging natural language supervision for learning is its scalability, unlike traditional crowd-sourced labeling which requires classic ML-compatible annotations. CLIP can passively learn from the abundant textual data available on the internet, establishing connections between visual representations and language, facilitating versatile zero-shot transfer across various tasks. The CLIP model (Figure~\ref{fig:CLIP} encodes text and image inputs separately using a text encoder and an image encoder, transforming them into embeddings in a joint multimodal space, where relevance scores are calculated to measure how well the image matches the text, useful for tasks like image-text retrieval or classification.

\section{Overview of Foundation Models}
\label{overviewofFM}
FMs revolutionized the way intelligent algorithms work by enabling a single model to perform various downstream tasks after training on large-scale unlabeled datasets \cite{wornow2023shaky}. The fundamental architecture of FMs relies on deep neural networks, particularly transformers trained on large-scale data. Below we provide a brief overview and describe the working mechanism of FMs.
\subsection{Learning architecture}
\label{learningArch}

The underlying architecture of FMs is typically based on transformers that can capture long-range dependencies and contextual relationships within data (Section~\ref{Attention}). The main architectures are encoder-decoder, encoder only, and decoder only, are explained below:

\subsubsection{Encoder-decoder}
The original transformer model \cite{vaswani2017attention} is based on encoder-decoder architecture. The encoder maps the input sequence of symbol representations $X=(x_1,x_2,\ldots,x_n)$ to a embedding space $z=(z_1,z_2,\ldots,z_n)$. Then the decoder works in an auto-regressive manner and generates the output $Y=(y_1,y_2,\ldots,y_n)$ one element at a time. This process involves utilizing the previously generated symbols as additional input at each step, allowing the model to factor in context from the ongoing generation \cite{vaswani2017attention}. Multiple FMs use encoder-decoder architecture, including BART \cite{lewis2019bart}, Segnet \cite{segnet}, and T5 (Text-to-Text Transfer Transformer) \cite{raffel2020exploring}.

\subsubsection{Encoder-only}
Encoder-only models encode contextual information from input sequences that extract and encode relevant rich and informative representations of the input sequence that can be used for various downstream tasks. Encoder-only models include BERT \cite{devlin2018bert}, ViT \cite{ViT}, and CLIP \cite{CLIP}.

\subsubsection{Decoder-only}
Decoder-only models specialize in the creative generation of coherent sequences. They predict the next token when given a sequence of tokens, which results in the generation of contextually meaningful outputs. Many SOTA models, such as BLOOM \cite{bloom}, PaLM \cite{palm}, and open pre-trained transformers \cite{opt}, are based on decoder-only architecture. GPT \cite{gpt} models, such as GPT-2 \cite{gpt2}, GPT-3 \cite{gpt3}, and the more recent GPT-4 \cite{gpt4}, are decoder-only models pre-trained on large-scale unsupervised text data.

\subsection{Learning algorithms}
SSL enables FMs to learn rich contextual representations from unlabeled data that can be fine-tuned for specific downstream tasks. SSL tackles the challenges associated with expensive, biased annotation and narrowly applicable supervised learning by providing a more robust and efficient approach for generalizing across downstream tasks. SSL methods that require supervised fine-tuning steps with manually labeled data have limited usability, especially for medical datasets, which are difficult to collect and annotate \cite{tiu2022expert}. Therefore, zero-shot learning-based methods are used instead, which do not require explicit manual or annotated labels during training \cite{tiu2022expert}. Zero-shot learning-based methods include the GPT models \cite{gpt,gpt2,gpt3,gpt4}, CLIP \cite{CLIP}, PaLM\cite{palm}, and SLIP \cite{SLIP}.

\subsection{Human feedback reinforcement learning}
FM models that learn from human feedback tend to demonstrate improved performance and can be generalized to other domains, unlike supervised learning \cite{stiennon2020learning}. For instance, Stiennon et al. \cite{stiennon2020learning} used the GPT-3 model with 6.7 billion parameters to predict human-preferred summaries from large-scale data and used the model acts as a reward function to refine a summarization policy through RL. Specifically, for Reddit posts \cite{volske2017tl}, multiple summaries were sampled from various sources. The summaries were then evaluated by human labelers and a reward model was trained to predict the likelihood that one summary was preferred over the other, based on human judgments.

\subsection{Prompting Foundation Models}

Prompting FMs are directed to perform a specific task to enhance their effectiveness and relevance \cite{awais2023foundational}.  In promoting FMs, users provide a prompt or a series of prompts that guide the model's response generation process. It aids the model to generate a response that is more aligned with the user's expectations. Incorporating additional context into the training and evaluation prompting FMs aids in improving the model's performance by providing relevant information and guiding its decision-making process.  For instance, GPT-based application \cite{gpt3, gpt4} possesses strong language understanding capabilities that can be prompted with specific textual inputs, such as questions. These can be further fine-tuned for tasks like sentiment analysis, text generation, or document summarization.  Furthermore, prompts can be used to specify the desired task such as image classification or caption generation that allows the model to focus its attention and resources accordingly. Models such as CLIP \cite{CLIP} which is trained on both image and text encoders using a contrastive pre-training objective rely on textual prompts for supervision. Furthermore, CLIP variants such as MaskCLIP \cite{dong2023maskclip} based on object detection and segmentation to align image regions with textual descriptions. 

\section{Flagship Foundation Models}
\label{falgship}
This section highlights the key FMs and their variants.

\subsection{BERT}
BERT \cite{devlin2018bert} was developed for language representation through training on a large corpus of textual data. It is a pre-trained bidirectional transformer model that utilizes contextual information within sequences from unlabeled data through the use of masked language modeling (MLM). The BERT frameworks consist of two main stages: pre-training and fine-tuning. The pretraining stage consists of two tasks: (i). MLM, in which the model randomly masks 15\% of input tokens and predicts these masked tokens. However, [MASK] tokens are not present during fine-tuning, causing a mismatch between pre-training and fine-tuning phases. To address this, the authors implemented a strategy during pre-training where [MASK] tokens were utilized 80\% of the time, a random token 10\%, and the unchanged token was used 10\% of the time during pre-training. The authors showed that existing techniques were unable to determine the relationship between multiple sentences and were only focused on individual sentences. Therefore, NSP was used to predict the next sentence that achieved over 97\% accuracy for the binary classification task (next sentence vs not next sentence).

During fine-tuning, a self-attention mechanism is used for downstream tasks with limited labeled data. Multiple variants of BERT models were then introduced such as RoBERTa (robustly optimized BERT pretraining approach) \cite{liu2019roberta} that surpassed the BERT mode. It is trained for longer (500K steps) on a large dataset of over 160 GB of English-language corpora. Furthermore, the [MASK] token was applied dynamically, creating a new masking pattern each time at the input to the model. In addition, the NSP loss module was removed.

Although BERT has been reasonably successful, its computation complexity and communication overhead are major limitations to adopting such models at scale \cite{albert}. Therefore, models inspired by BERT were developed to achieve BERT-level performance while reducing the training time and model size. For instance, Sanh \cite{distilbert} proposed the lightweight DistilBERT. It is based on the distillation knowledge compression technique \cite{hinton}, which is a smaller model (student) and trained to reproduce a larger model (teacher). The number of layers in DistilBERT is reduced by half and the architecture has been optimized \cite{distilbert}. 
The proposed model was nearly 40\% smaller in size and 60\% faster than BERT with comparable performance on various downstream tasks.

A Lite BERT (ALBERT) \cite{albert} was 1.7 times faster, used 18 times fewer parameters, and achieved better performance than BERT. A model based on a denoising autoencoder known as BART \cite{lewis2019bart} has been developed for sequence-to-sequence modeling and shows promising performance. However, it is computationally expensive, with 10\% more parameters than BERT.

\subsection{GPT}
The ChatGPT we interact with today originated from extensive research trials based on a GPT model.
Radford et al. \cite{gpt} proposed a transformer-based GPT model that utilizes an unsupervised approach during pre-training and supervised fine-tuning for natural language understanding tasks. GPT was trained on the BooksCorpus dataset \cite{zhu2015aligning} for unsupervised pre-training. Then, the parameters learned at the pre-training stage were used for a target task with a manually annotated dataset in a supervised manner. A separate work by Radford et al. \cite{gpt2} presented a GPT-2 language model with a zero-shot setting for multiple downstream tasks without the need for any major modifications. GPT-2 was trained on a large amount of data --- WebText --- contains over 8 million cleaned documents with more than 40 GB of text. GPT2 performed well on multiple tasks, including reading comprehension, summarization, translation, and question-answering. However, although it performed better than baselines, GPT-2 still could not be applied practically as the performance was still not promising. Therefore, a more robust model, GPT-3, with 175B parameters, was presented and performed well on various downstream tasks. 
It was trained on 300 billion tokens (570 GB) of plain compressed filtered text. The performance of GPT-3 for various downstream tasks surpasses that of existing techniques when tested using zero-shot, one-shot, and few-shot learning scenarios. One key solution to reduce untruthful or irrelevant outputs is to incorporate human feedback, such as in InstructGPT \cite{InstructGPT}. InstructGPT uses reinforcement learning from human feedback (RLHF) to fine-tune the GPT-3 model on human data and reduce the number of parameters to only 1.3B. Another more sophisticated language model known as ChatGPT \footnote{\url{https://chat.openai.com/}} or GPT3.5 provides more user-friendly, interactive, and context-aware responses and conversational experience. It follows a similar refined architecture to that of InstructGPT \cite{InstructGPT}.
ChatGPT is limited to text inputs, but this limitation was addressed in the GPT-4 \cite{gpt4} model, which accepts images in addition to text inputs while producing text outputs. However, the details of GPT4 are not publicly available and a major issue with such language models is hallucinations \cite{gpt4}.

\subsection{DALL-E}

DALL-E \cite{DALLE} is a transformer-based text-to-image generative model with a two-stage training procedure. In the first stage, each RGB image is compressed into a 32$\times$32 grid of image tokens, bypassing the direct use of pixels, significantly reducing memory consumption. Furthermore, short-range dependencies between pixels are prioritized during training. Therefore, in the second stage, an autoregressive transformer models the joint distribution over text and image tokens. The proposed model was trained on a dataset of 250 million text-image pairs gathered from online sources.
An extension DALL-E-2 \cite{DALLE2} has also been presented, which can generate high-resolution images based on text prompts and existing images. It uses CLIP-based robust representation to enhance its image generation. DALL-E-2 first generates an image embedding induced by the text input using CLIP. Then, this generated image embedding is used as an input of the diffusion model to create an image. The model is trained using an image ($x$)-caption ($y$) pair. 

\subsection{Llama}
LLaMa \cite{llama} is trained on 1.4 trillion tokens and is based on existing large pre-trained models with key modifications. For instance, pre-normalization from GPT 3 was adopted to improve its training stability, but unlike in GPT-3, normalization was performed at the input of each transformer sub-layer. The SwiGLU \cite{SwiGLU} activation function from the PaLM model \cite{palm} replaced the ReLU non-linearity activation function. The absolute positional embeddings were replaced with rotary positional embeddings (RoPE) \cite{su2024roformer} from the PaLM model \cite{palm} at each layer. 
LLaMA has 65B parameters several times more than GPT-3, and performed well in various downstream tasks. A more refined version called LLaMA 2 \cite{llama2}, with 70B parameters in addition to Llama 2-Chat, is designed for dialogue applications and performance surpasses open-source chat models on various benchmarks \cite{llama2}.

\subsection{LLaVA}

Large Language and Vision Assistant) \cite{liu2024visual} is a large multimodal model designed to integrate vision and language for general-purpose tasks such as image understanding, captioning, and reasoning.  LLaVa was built by instruction-tuning a language model using machine-generated multimodal (image-text) instruction-following data created by GPT-4. LLaVA connects a CLIP visual encoder with the Vicuna language model, fine-tuned on approximately 158,000 image-text pairs to develop its instruction-following abilities. The training process is split into two stages: first, pre-training aligns image features with the language model's tokens, and second, end-to-end fine-tuning integrates image-text pairs to improve the model's instruction-following capability. The model exhibits strong multimodal reasoning capabilities, achieving high performance on tasks like ScienceQA and performing well even in out-of-domain scenarios.  Although LLaVA demonstrates promising performance, it struggles to interpret medical images due to complex visual patterns accurately, often resulting in incomplete or inaccurate responses. Therefore, tailored for healthcare data, LLaVA-Med \cite{li2024llava}was introduced. LLaVA-Med was trained using a novel curriculum learning approach on a large PMC-15M dataset containing 15 million biomedical image-text pairs extracted from PubMed Central. By using GPT-4 to generate instruction-following data from these pairs, the model was trained to handle open-ended conversational tasks about biomedical images. It outperformed previous state-of-the-art models on several benchmarks, such as biomedical visual question answering, highlighting its utility in clinical settings. Other models inspired by LLaVA include \cite{lin2024moe,sun2024dr,li2024llavaS,jin2024surgical,seyfioglu2024quilt}

\subsection{Stable Diffussion}
Diffusion models (DMs) \cite{ho2020denoising} can synthesize high-quality images but demand significant computational resources during training. Furthermore, they require costly evaluations in pixel space. Therefore, an efficient model called the stable diffusion model \cite{stabledifuusion} explicitly separates the compressive and generative learning phases. The proposed model employs an autoencoder that generates a lower-dimensional representation space and is computationally efficient. It enhances the computational efficiency of sampling, leveraging the inductive bias of UNet-based diffusion models for spatially structured data and providing a multipurpose compression model applicable to various generative tasks. The proposed diffusion model demonstrates competitive performance in tasks such as image synthesis, inpainting, and super-resolution and reduces computational costs compared to pixel-based DMs. An improved version of the stable diffusion model, known as SDXL \cite{sdxl}, incorporates a larger UNet backbone (2.6B parameters in the UNet), additional attention blocks, and a broader cross-attention context. 

\subsection{Industrial pretrained models}

Industrial-scale pre-trained models are extensively employed across a spectrum of industrial applications. Here we briefly mention some SOTA models widely adopted in various industries including healthcare.
Language Models for Dialog Applications (LaMDA) \cite{lamda} is built on a transformer architecture that contains 137B parameters tailored for dialog applications. It addresses two challenges faced by language models: safety and factual grounding. The safety enhancements involve filtering responses using a classifier fine-tuned with crowd worker-annotated data, aligning model behavior with human values based on Google AI principles\footnote{\url{https://ai.google/responsibility/principles/}}. Factual grounding enables the model to reference external knowledge sources to ensure responses are grounded in verifiable information. 

A more robust model than GPT-3 known as GLaM (Generalist Language Model) \cite{glam} has been proposed by Google, which has 1.2 trillion parameters. GLaM achieved better performance on 29 NLP tasks in nearly all zero-shot, one-shot, and few-shot learning scenarios, consuming only one-third of the energy of GPT-3.
A more refined version, the Pathways Language Model (PaLM) \cite{palm}, with 540 billion parameters, has also been presented by Google. It is a decoder-only-style transformer model with multiple modifications, such as SwiGLU activations \cite{SwiGLU} for MLP intermediate layers, parallel formulation for a faster training speed, multi-query attention, and RoPE embeddings \cite{su2024roformer} for improved performance on long sequences. PaLM has achieved SOTA performance in reasoning tasks and shows proficiency in multilingual tasks and source code generation. A more efficient model, PaLM 2 \cite{palm2}, was recently developed by Google and performs better than PaLM on various downstream tasks, with enhanced multilingual, reasoning, and coding capabilities. It is utilized at various sizes in 25 Google products, including Bard \footnote{\url{https://bard.google.com/}}, the health-focused Med-PaLM \cite{medpalm}, Med-Plam 2 \cite{medpalm2}, and the security-focused Sec-PaLM\footnote{\url{https://cloud.google.com/security/ai/}}.  
SenseNova\footnote{\url{https://www.sensetime.com/en/news-detail/51166397?categoryId=1072}}, developed by SenseTime, is a large-scale FM that offers various API interfaces and services specifically designed for enterprise customers and widely adopted in different industries, mainly in China, such as in the automotive sector, where SenseNova was instrumental in the creation of the BEV (Bird's-Eye-View) system that can recognize 3,000 types of objects. Furthermore, SenseNova introduced the ``SenseChat''\footnote{\url{https://chat.sensetime.com/wb/login}} model, which is similar to ChatGPT but specifically designed for use in China.

\footnotesize 
\begin{table}[htbp]
\centering
\caption{Flagship Foundation Models. E-WP: English Wikipedia, CC: Common Crawl, WT: WebText. The link is provided if the models publicly available.}
\label{tab:models}
\footnotesize
\begin{tabular}{p{0.5cm}p{0.5cm}p{1cm}p{0.9cm}p{2cm}p{0.5cm}p{1cm}}
\hline
\textbf{Work} & \textbf{Year} & \textbf{Model} & \textbf{Parameter} & \textbf{Training Datasets} & \textbf{Public} & \textbf{Task} \\
\hline
    \cite{devlin2018bert} & 2018 &  BERT &  340M & \cite{zhu2015aligning} , E-WP 
    &   \href{https://github.com/google-research/bert} \cmark 
    & NLP  \\
    
    \cite{liu2019roberta} & 2019 &  RoBERTa &  355M &  \cite{zhu2015aligning}, E-WP, CC-News, Open WT \cite{Gokaslan2019OpenWeb}, stories \cite{trinh2018simple}  
    &   \href{https://github.com/facebookresearch/fairseq/blob/main/examples/roberta/README.md} \cmark 
    & NLP \\

    \cite{distilbert} & 2019 &  DistilBERT & 66M & \cite{zhu2015aligning}, E-WP 
    &    \href{https://github.com/huggingface/transformers} \cmark 
    & NLP \\
    \cite{albert} & 2020 &  ALBERT &  235M & E-WP,\cite{zhu2015aligning}   
    &    \href{https://github.com/google-research/ALBERT}\cmark 
    & NLP \\

    \cite{gpt} & 2018 &  GPT &  117M &  \cite{zhu2015aligning} 
    &  \href{https://github.com/openai/finetune-transformer-lm}\cmark 
    & NLP \\
    
    \cite{gpt2} & 2019 &  GPT-2 &  1.5B &  WT
    &  \href{https://github.com/openai/gpt-2}\cmark 
    & NLP \\
    
    \cite{gpt3} & 2020 &  GPT-3 &  175B &  CC (filtered), WT2, Books1, Books2, E-WP & \xmark   & NLP \\
    \cite{gpt4} & 2023 &  GPT-4 &  - &   & \xmark    & NLP and image \\
    \cite{InstructGPT} & 2022 & InstructGPT &  1.3B &  CC (filtered), WT2, Books1, Books2, E-WP &  \xmark & NLP \\
    
    \cite{DALLE} & 2021 & DALL-E &  12B&  250M image-text pairs from internet 
    & \href{https://github.com/openai/DALL-E}\cmark 
    & text-to-image synthesis \\
    
    \cite{DALLE2} & 2022 &  DALL-E &  3.5B & 650M image-text pairs  
    &  \href{https://github.com/lucidrains/DALLE2-pytorch}\cmark
    & text-to-image synthesis \\
    
    \cite{llama} & 2023 &  LLaMA &  65B &  CC, C4 \cite{raffel2020exploring}, Github, E-WP, Books, ArXiv, StackExchange
    &  \href{https://github.com/meta-llama/llama}\cmark 
    & NLP \\
    
    \cite{llama2} & 2023 &  LLaMA2 &  70B &  2 trillion tokens public data 
    &\href{https://www.llama.com/llama2/}\cmark 
    & NLP \\
    
    \cite{stabledifuusion} & 2022  &  Stable diffusion &  1.45B &  LAION \cite{schuhmann2021laion} 
    & \href{https://github.com/CompVis/latent-diffusion}\cmark 
    & Image synthesis \\
    
    \cite{sdxl} & 2023 &  SDXL &  3.5B &  - & 
    \href{https://github.com/Stability-AI/generative-models}\cmark 
    & Image synthesis \\
    
    \cite{lamda} & 2022 &  LaMDA & 1.2T &  Dialog data, C4
    & \href{https://github.com/conceptofmind/LaMDA-rlhf-pytorch}\cmark
    & Dialog Application \\
    
    \cite{glam} & 2022 & GLaM & 137B & Filtered Webpages, E-WP, Conversations, Forums, Books, News & \xmark  &
NLP tasks\\
    \cite{palm} & 2022 &  PaLM &  540b &  780 billion tokens & \xmark & NLP \\
    \cite{palm2} & 2023  & PaLM2 & 13B &   & \xmark  & NLP  \\
    \cite{sensetime}   & 2023 &  SenseNova &  - &  - & \xmark  &  NLP \\
\hline
\end{tabular}
\end{table}

\ifCLASSOPTIONcaptionsoff
  \newpage
\fi

\bibliographystyle{IEEEtran}

\bibliography{bibtex/bib/IEEEexample}


\end{document}